\documentclass{article}

\PassOptionsToPackage{numbers, compress}{natbib}
\usepackage[final]{neurips_2021}

\usepackage[utf8]{inputenc} %
\usepackage[T1]{fontenc}    %
\usepackage{hyperref}       %
\usepackage{url}            %
\usepackage{booktabs}       %
\usepackage{amsfonts}       %
\usepackage{nicefrac}       %
\usepackage{microtype}      %
\usepackage[dvipsnames]{xcolor}  %
\usepackage{multirow}
\usepackage{graphicx}
\usepackage[normalem]{ulem}

\usepackage[mathcal]{euscript}
\usepackage{wrapfig}
\usepackage{dsfont}
\usepackage{amsmath,xfrac,amsthm}
\usepackage{amssymb}
\usepackage{bm}
\newtheorem{innercustomthm}{Theorem}
\newenvironment{customthm}[1]
  {\renewcommand\theinnercustomthm{#1}\innercustomthm}
  {\endinnercustomthm}
\renewcommand{\paragraph}[1]{\textbf{#1}}
\usepackage[ruled,vlined]{algorithm2e}

\usepackage{xr}
\makeatletter
\newcommand*{\addFileDependency}[1]{%
  \typeout{(#1)}
  \@addtofilelist{#1}
  \IfFileExists{#1}{}{\typeout{No file #1.}}
}
\makeatother

\newcommand{\cut}[1]{\mathrm{cut}\hspace{-1px}\left(#1\right)}
\newcommand{\entropy}{\textrm{H}}

\DeclareMathOperator{\length}{\mathrm{L}}
\DeclareMathOperator{\E}{\mathbb{E}}

\usepackage{mathtools}

\newcommand{\univ}{\mathfrak{U}}
\renewcommand{\b}[1]{\mathbf{#1}}

\newcommand{\bx}{\b{x}}
\newcommand{\bX}{\b{X}}
\newcommand{\bh}{\b{h}}

\newcommand{\cG}{\mathcal{G}}
\newcommand{\cH}{\mathcal{H}}

\newcommand{\cX}{\mathcal{X}}

\newcommand{\cN}{\mathcal{N}}

\usepackage{mathtools}

\newtheorem{theorem}{Theorem}

\newtheorem{lemma}{Lemma}

\title{Partition and Code: learning how to compress graphs}

\author{%
Giorgos Bouritsas\thanks{This work was done while Giorgos Bouritsas was visiting Dr. Andreas Loukas at EPFL, Lausanne.}\\
Imperial College London, 
UK\\
\texttt{g.bouritsas@imperial.ac.uk}

\And
\hspace{-0.1cm}Andreas Loukas\\
\hspace{-0.1cm}EPFL, 
Switzerland\\
\hspace{-0.1cm}\texttt{andreas.loukas@epfl.ch}

\AND
\hspace{0.7cm}Nikolaos Karalias \\
\hspace{0.8cm}EPFL, 
Switzerland\\
\hspace{0.7cm}\texttt{nikolaos.karalias@epfl.ch} \\

\And
Michael M. Bronstein \\
Imperial College London / Twitter,
UK\\
\texttt{m.bronstein@imperial.ac.uk} \\
}

\begin{document}

\maketitle
\setcounter{footnote}{0}

\begin{abstract}
Can we use machine learning to compress graph data? The absence of ordering in graphs poses a significant challenge to conventional compression algorithms, limiting their attainable gains as well as their ability to discover relevant patterns. On the other hand, most graph compression approaches rely on domain-dependent handcrafted representations and cannot adapt to different underlying graph distributions. This work aims to establish the necessary principles a lossless graph compression method should follow to approach the entropy storage lower bound. Instead of making rigid assumptions about the graph distribution, we formulate the compressor as a probabilistic model that can be learned from data and generalise to unseen instances. Our ``\textit{Partition and Code}'' framework entails three steps: first, a  partitioning algorithm decomposes the graph into subgraphs, then these are mapped to the elements of a small dictionary on which we learn a probability distribution, and finally, an entropy encoder translates the representation into bits. All the components (partitioning, dictionary and distribution) are parametric and can be trained with gradient descent. We theoretically compare the compression quality of several graph encodings and prove, under mild conditions, that PnC achieves compression gains that grow either linearly or quadratically with the number of vertices. Empirically, PnC yields significant compression improvements on diverse real-world networks.\footnote{The source code is publicly available at \url{https://github.com/gbouritsas/PnC}}
\end{abstract}

\section{Introduction}
\label{sec:intro}

Lossless data compression has been one of the most fundamental and long-standing problems in computer science.  It is by now well-understood that the intrinsic limits of compression are governed by the entropy of the underlying data distribution \cite{shannon1948mathematical}. 
Crucially, these limits expose an intimate connection between compressibility and machine learning: the better one models %
the underlying data distribution (from limited observations) the more bits can be saved and vice-versa \cite{mackay2003information}. 

The compression of ordered data such as text, images, or video, underpins the modern technology from web protocols to video streaming. However, graph-structured data remain a notable exception. As graph data are becoming more  prevalent, it becomes increasingly important to invent practical ways to encode them parsimoniously.

There are three main challenges one faces when attempting to compress graphs:

\emph{C1. Dealing with graph isomorphism (GI).} A key difficulty that distinguishes graphs from conventional data lies in the absence of an inherent ordering of the graph vertices. In order to be able to approach the storage lower bounds, isomorphic graphs should be encoded with the same codeword. %
However, since the complexity of known algorithms for GI is super-polynomial on the number of vertices \cite{babai2016graph}, a direct use of GI is impractical for graphs consisting of more than a few hundred vertices.
Indeed, an examination of the graph compression literature reveals that most progress has been made by optimising a vertex ordering and adapting methods originally invented for vectored data \cite{vigna2004webgraph, lim2014slashburn, dhulipala2016compressing}. Unfortunately, naively encoding graphs as vectors results in a significant loss in compression\footnote{This follows by a simple counting argument: there are $2^{n \choose 2}$ labelled undirected graphs while the respective number for unlabelled graphs is asymptotically equal to $2^{n \choose 2}/n!$~\cite{harary2014graphical}. Thus, if all graphs of $n$ vertices are equally probable, an encoding that does not consider isomorphism would sacrifice $\log{n!}$ bits~\cite{naor1990succinct, choi2012compression}.}. %

\textit{C2. Evaluating the likelihood.} %
An optimal encoder~\cite{shannon1948mathematical} requires one to accurately estimate and evaluate the probabilities of all the possible outcomes of the underlying domain. When dealing with high-dimensional data, this can be addressed by partitioning the data into parts to obtain a decomposition of the probability distribution: e.g., images can be compressed by modelling the distribution of pixels or patches \cite{sneyers2016flif, van2016pixel, mentzer2019practical}, and text by focusing on characters or n-grams \cite{welch1984technique, ziv1977universal,ziv1978compression,bell1989modeling,burrows1994block,schmidhuber1996sequential, mahoney2000fast}. However, since graphs do not admit an efficiently computable canonical ordering, it is unclear what decomposition one should employ. %

\textit{C3. Accounting for the description length of the learned model.} The classical learning theory trade-off between model complexity and generalisation %
is of paramount concern for effective compression. Though in typical deep learning applications one can aim to model the data distribution with an overparametrised neural network (NN) that generalises well, utilising such models to compress information is problematic: since decoding is impossible unless the decoder also receives the learned model (i.e., the NN parameters), \emph{overparametrised models are, by definition, suboptimal}. This is a pertinent issue for likelihood-based neural approaches as overparametrisation is commonly argued to be a key component of why NNs can be trained~\cite{du2018gradient, arora2018optimization,allen2019convergence}.

\begin{wrapfigure}{R}{0.44\textwidth}
      \includegraphics[width=0.46\textwidth]{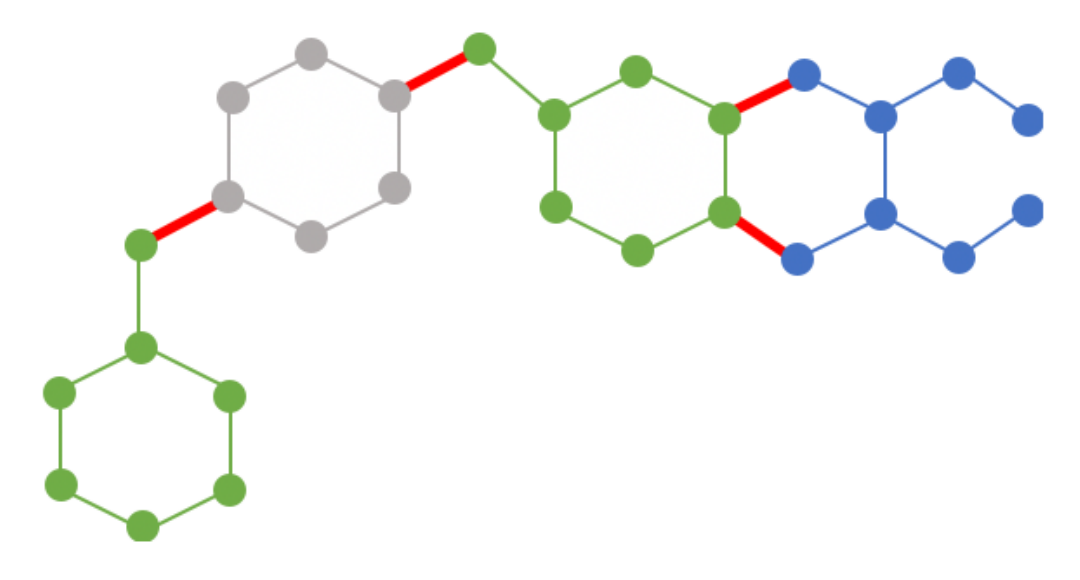}
    \caption{Illustration of the graph decomposition. The subgraph colours  correspond to dictionary atoms \textcolor{gray}{$a_1$}, \textcolor{LimeGreen}{$a_2$} and \textcolor{RoyalBlue}{$a_3$}. Cuts are denoted in red.}
     \label{fig:method_illustration}
\end{wrapfigure}

\paragraph{The Partition and Code (PnC) framework.}
Our main contribution is PnC, a framework for learning compression algorithms suitable for encoding graphs sampled from an underlying distribution. %
In the heart of PnC lie two ideas: 
\textit{(a) Learn to break the problem into parts.} Rather than predicting directly the likelihood, we aim to learn how to decompose graphs into non-overlapping {subgraphs},
see Fig. \ref{fig:method_illustration}. 
\textit{(b) Identify and code recurring subgraphs when possible.} We use a learned \textit{dictionary} to code subgraphs that appear frequently, whereas rare subgraphs are encoded separately. %
The dictionary is restricted to contain only a finite number of small recurring subgraphs.
This biases the model towards interpretable and well generalising solutions. 
Both the ``partition'' and ``code'' components of PnC are learned directly from the data, by optimising the total description length. %

Our framework provides a solution to the three challenges of graph compression. C1: By constraining the dictionary to only contain graphs of size up to a small constant, we can efficiently solve GI.
C2: Graph partitioning provides us with the desirable decomposition of the distribution. Using appropriate parametrised probabilistic models, we obtain a
closed-form expression for the likelihood that can be later used by any close-to-optimal entropy coder~\cite{huffman1952method,witten1987arithmetic, duda2013asymmetric}. 
C3: Since we use NNs to \textit{decompose} the distribution (and not to predict the likelihood), we can rely on overparametrisation without having to relay the NN parameters to the decoder. Also, the complexity of our learned hypothesis (the dictionary) can be computed, and thus optimised, during training. 

\textbf{Theoretical results.} Our analysis reveals that PnC can significantly improve upon less sophisticated graph encoders and justifies the usefulness of both the ``Partition'' and the ``Code'' component.
Specifically, we %
prove that under mild conditions on the underlying graph distribution, PnC requires in expectation $\Theta(n^2)$ less bits than standard graph encodings, even if the latter are given access to an oracle that solves GI. %
Further, the dictionary induces additional savings of $\Theta(n)$ bits, with the gain being inversely proportional to the entropy of the distribution of the dictionary atoms. Thus, the more repetitive the patterns in the graph distribution are, the larger will be the compression benefits of PnC.

\textbf{Practical algorithms.} We instantiate of our framework using the following algorithmic modules: 
(a) a low-parameter learnable estimate of the probability distribution,
(b) learning to select the dictionary from a
graph universe of finite size, %
and (c) learning to partition. The latter is a parametric randomised iterative algorithm, the probabilities of which are inferred from a Graph Neural Network (GNN) and optimised with reinforcement learning. Importantly, all algorithms can be jointly trained in order to minimise the total description length in a synergistic manner. We evaluate our framework on diverse real-world graph distributions and showcase compression gains with respect to both conventional and advanced baseline compressors,  in observed and unseen data.

\section{Related work}

\textbf{Engineered codecs.} 
The majority of graph compressors are not probabilistic, but
rely on hand-engineered encodings optimised to take advantage of  domain-specific properties of e.g., WebGraphs~\cite{vigna2004webgraph}, social networks~\cite{dhulipala2016compressing, apostolico2009graph, boldi2011layered}, and biological networks~\cite{peshkin2007structure}. A common idea in these approaches is \textit{vertex reordering} \cite{vigna2004webgraph, lim2014slashburn, dhulipala2016compressing, apostolico2009graph,boldi2011layered, chierichetti2009compressing, claude2010fast, rossi2018graphzip}, where the adjacency matrix is permuted in such a way that makes it ``compression-friendly'' for mainstream compressors of sequences, such as $\mathsf{gzip}$. The algorithms identifying the re-orderings are usually based on heuristics taking advantage of specific network properties, e.g., community structure. Another recurrent idea is to detect or use predefined \textit{frequent substructures} (e.g., cliques) to represent more efficiently different parts of the graph via grammar rules \cite{maneth2016compressing}. These approaches do not attempt to model the underlying graph distribution and thus to approach the storage lower bounds, but strive to find a balance between compression ratios and fast operations on the compressed graphs. Thus, despite their practical importance, they are less relevant to our work. A comprehensive survey can be found in \cite{besta2018survey}.

\textbf{Theory-driven approaches.} 
Several works have contributed to the foundations of the information content and the complexity of graphs~\cite{naor1990succinct, choi2012compression, trucco1956note, trucco1956information, korner1973coding, turan1984succinct, dehmer2008information, dehmer2009large, mowshowitz2012entropy}. However, few works have attempted to model the underlying graph distribution.
Perhaps the most outstanding progress has been made for graphs modelled by the Stochastic Block Model (SBM)~\cite{holland1983stochastic,rosvall2008maps,karrer2011stochastic, peixoto2012entropy,peixoto2013parsimonious,peixoto2017nonparametric,peixoto2019bayesian,lee2019review}. Although originally invented for clustering and network analysis purposes, these approaches can be seamlessly used for compression due to their exact likelihood computation. In fact, as we argue in this work, virtually any graph clustering algorithm can be used successfully for compression, by defining codewords corresponding to a community-based random graph model. However, as our experiments confirm, such approaches are less effective at compressing graphs that do not contain clusters.

\textbf{Likelihood-based neural approaches.}
Any generative model that can provide likelihood estimates in a finite sample space can be used for lossless compression. As a result, a plethora of likelihood-based neural compressors have been recently invented, ranging from autoregressive models for text \cite{schmidhuber1996sequential, mahoney2000fast, cox2016syntactically, goyal2018deepzip} and images \cite{van2016pixel, mentzer2019practical} to latent variable models \cite{DBLP:conf/iclr/TownsendBB19, kingma2019bit, TownsendBKB20, ruan2021improving, severo2021compressing} (paired with bits-back coding \cite{wallace1990classification, hinton1993keeping} and Asymmetric Numeral Systems - ANS \cite{duda2013asymmetric}), normalising flows \cite{HoLA19, DBLP:conf/nips/HoogeboomPBW19, TranVADP19, van2020idf++}  and most recently, diffusion-based generative models \cite{kingma2021variational, hoogeboom2021autoregressive}.  However, 
the vast majority of current graph generators lack the necessary theoretical properties an effective graph compressor should have: they compute probabilities on labelled graphs instead of isomorphism classes by resorting to a heuristic ordering ~\cite{JinBJ18, li2018learning, you2018graphrnn, LiuABG18, liao2019efficient, dai2020scalable, ShiXZZZT20, LuoYJ21} (in general this will be suboptimal unless we canonicalise the graph/solve graph isomorphism, while different orderings will have non-zero probabilities,
hence we will incur compression losses),
and/or do not provide a likelihood~\cite{de2018molgan, yang2019conditional, bojchevski2018netgan, niu2020permutation, liu2021graphebm}.  

An important caveat, which is often ignored in the literature, is that when parametrising the distribution with a neural network,  the data cannot be recovered unless the decoder has access to the network itself. Hence it is necessary to account for the NN's description length when evaluating compression gains.  However, generative models are usually parameter inefficient, while compressing them (especially during training) is a challenging problem \cite{bird2021reducing}. Note that this is significantly different than compressing neural classifiers, since the capacity to infer the likelihood up to high precision needs to be retained. Therefore, these approaches can have diminishing (or even negative) returns when the dataset size is not large enough. In contrast, by optimising the total description length (equivalent to \textit{maximum a-posteriori}), we design a compressor that is practical even for small datasets, while the learned dictionary makes our compressor interpretable.

Other related work includes using  compression objectives paired with heuristic algorithms for downstream tasks, such as motif finding \cite{ketkar2005subdue, bloem2020large} and graph summarisation \cite{koutra2014vog, lefevre2010grass},  lossy compression/coarsening \cite{nourbakhsh2015matrix,loukas2018spectrally, loukas2019graph, GargJ19, dong2020copt, pmlr-v108-jin20a}, and graph dictionary learning in the context of sparse coding \cite{zhang2012learning,Vincent-CuazVFC21}.

\section{Preliminaries}\label{sec:prelims}

We use capital letters for sets $A,B$, bold font for vectors $\bx$ and matrices $\bX$, and  calligraphic font $\cX$ for families of sets. We  express the information content in bits by using a base-2 logarithm $\log$.

\noindent\textbf{Graphs.} 
Let $G = \{V, E\}$ 
be a graph with $n$ vertices $V = \{v_1, \ldots, v_n \}$ and $m$ edges, where $e_{ij} \in E$ whenever vertices $v_i,v_j \in V$ are connected by an edge. For simplicity, we assume the graphs to be undirected, though the same methods apply to directed graphs as well with minimal modifications. Let $\cut{A, B} = \{ e_{ij} \in E \ | \ v_i \in A \text{ and } v_j \in B\}$ denote the set of edges with endpoints at two different vertex sets $A, B \subseteq V$. 

\noindent\textbf{Information theory.}
Following the usual terminology in information theory and Minimum Description Length (MDL) theory \cite{mackay2003information, grunwald2007minimum, cover1999elements}, we assume an {\em observation space} $\mathfrak{G}$ (in our case a space of possible graphs), and a probability distribution $p$ (sometimes referred to as {\em probabilistic source}) producing samples from the observation space. We observe a dataset  $\cG = \{G_1, G_2, \dots, G_{|\cG|}\}$ of i.i.d. observations drawn from $p$.
Note that this setting is in contrast with most works on graph compression \cite{vigna2004webgraph, claude2010fast, lim2014slashburn, dhulipala2016compressing}, where the target is to compress a {\em single} large network, such as a social network or a web graph.
Let a \textit{description method} or \textit{symbol code} $\mathsf{CODE}: \mathfrak{G} \to \{0,1\}^*$ be a mapping from the observation space to a variable-length sequence of binary symbols, the output of which is a \textit{codeword}.

In the context of graph compression, we are interested in the description length of the code $\length_\mathsf{CODE}(G)$, or $\length(G)$ for brevity, i.e., the number of bits needed to encode the graph $G$, rather than the code itself, given a single requirement: the code needs to be \textit{uniquely decodable}, meaning that any concatenation of codes can be uniquely mapped to a sequence of observations. This property can be easily verified using only the description lengths by the well-known \textit{Kraft–McMillan inequality} (l.h.s. in formula~(\ref{eq:kraft-inequality})), an important implication of which is that every uniquely decodable code implies a probability distribution $q(G)$ (r.h.s):
\begin{equation}\label{eq:kraft-inequality}
    \sum_{G \in \mathfrak{G}} 2^{-\length(G)} \leq 1 \ \  (\text{code}) \   \Longleftrightarrow \ 
    q(G) = \frac{2^{-\length(G)}}{\sum_{G \in \mathfrak{G}} 2^{-\length(G)}} \ \  (\text{distribution})
\end{equation}
This equivalence allows us to always define underlying probabilistic models when working directly with description lengths.
When the Kraft–McMillan inequality holds with equality, we say that the code is \textit{complete}; the case of strict inequality means the code is {\em redundant}. Moreover, in the case of complete codes, the expected length of the code corresponds to the distribution entropy $\mathbb{H}_q[G]$ (plus a constant term when the code is redundant). Thus, compression is synonymous to defining probabilistic models with the lowest possible entropy. An important notion that frequently appears in our theoretical analysis is the \textit{binary} entropy, i.e., the entropy of a bernoulli variable with probability of success $p$. To distinguish it from the general notion of entropy we will denote it as ${\entropy(p) = -p\log p - (1-p)\log(1-p)}$. It holds that $0\leq \entropy(p)\leq 1$, where the l.h.s equality is satisfied for $p \in \{0,1\}$ and the r.h.s. for $p=1/2$.

\noindent\textbf{Common graph encodings.}
An important principle that we follow is that whenever we cannot make any assumptions about an underlying distribution, or modelling it is impractical, then the uniform distribution $p_{\text{unif}}$ is chosen for encoding. The reason is that uniform distribution is \textit{worst-case optimal}~\cite{grunwald2007minimum}: for any non-uniform unknown distribution $p$, there always exists a distribution $q$ the corresponding encoding of which will be on expectation worse than the uniform encoding: $\E_{x\sim p }[-\log q(x)] > \E_{x\sim p }[-\log p_{\text{unif}}(x)]$. 
In the context of graphs, when we cannot make any assumptions or when enumeration is impossible, we will be using a slightly more informative distribution: the Erdős–Rényi (ER) random graph model. This model  assigns equal probability to all labelled graphs with $n$ vertices and $m$ edges. Assuming a uniform probability over the possible number of vertices $n$ and number of edges $m$ given $n$, we get:
\begin{align}\label{eq:null}
L_{\text{null}}(G) = \log (n_{\text{max}}+1) + \log \bigg({n \choose 2} + 1\bigg) + \log {{n \choose 2}  \choose m},
\end{align}
where $n_{\text{max}}$ is an upper bound on the number of vertices. The ``\textit{null}'' encoding compresses more efficiently  graphs that are either very sparse or very dense (low-surprise) as the number of possible graphs with $m$ edges is maximised when ${m = n(n-1)/4}$. %

\section{The Partition \& Code (PnC) graph compression framework} 

Our pipeline consists of three main modules: a \textit{partitioning} module, a \textit{dictionary} module, and an \textit{entropy encoding} module. (a) The partitioning module is responsible for decomposing the graph into disjoint subgraphs and cross-subgraph edges (or \textit{cuts}). Subgraphs play the role of elementary structures, akin to characters or words in text compression, and pixels or patches in image compression. (b) The dictionary is a small collection of subgraphs (\textit{atoms}) that are recurrent in the graph distribution. The dictionary module maps the partitioned subgraphs to atoms in the dictionary, allowing us to represent the graph as a collection of atom indices and cuts.  (c) Finally, this representation is given as input to the an entropy encoder that translates it into bits. 

Decoding the graph in a lossless way 
involves inverting these three steps: initially the atom indices and cuts are decoded using the same probabilistic model with the encoder, then the atoms are retrieved from the dictionary, and finally all elements are composed back to obtain the original structure. The composition becomes possible by making sure that the cuts are encoded w.r.t. an arbitrarily chosen ordering of each atom's vertices, hence the decoded graph is guaranteed to be isomorphic to the input, but not necessarily with the same vertex ordering.

\subsection{Step 1. Partitioning}\label{sec:decomp}
\label{subsec:partition}

The first step of PnC is to employ a parametric partitioning algorithm $\mathsf{PART}_\theta$ to decompose each graph $G$ into $b$ subgraphs of bounded size:
\begin{align}
    \mathsf{PART}_\theta(G) = (\cH, C) 
    \quad \text{and} \quad 
    \cH = \{H_1, H_2, \cdots, H_b\},
    \label{eq:partition}
\end{align}
where  
$ {H_i = \{V_i, E_i\}}$
is the $i$-th subgraph and
$C = \{V, E_C\}$ 
is a $b$-partite graph containing all cut edges $E_C = E - \cup_i E_i$. Variable $\theta$ indicates the learnable parameters.

 \textit{Is partitioning necessary?} In order to convert probability estimates to codewords, entropy encoders \cite{huffman1952method,witten1987arithmetic,duda2013asymmetric} need to be able to compute the probability of \textit{every graph in the observation space} (or equivalently to have access to the cumulative distribution function (c.d.f.)) - see challenge C2 in Section~\ref{sec:intro}. To achieve this goal, one needs to partition the observation space in  a way that permits efficient enumeration of the possible outcomes. Analogously, as we will see in Section~\ref{sec:encoding}, graph partitioning allows us to decompose the distribution and thus to obtain a closed-form expression for the likelihood. Further, Theorem \ref{thrm:main} suggests that partitioning brings a useful inductive bias for graph data providing significant storage gains compared to distribution-agnostic baselines.

\subsection{Step 2. Graph dictionary}

Rather than naively compressing each subgraph in~\eqref{eq:partition} under a null model, an effective compression algorithm should exploit regularities in the output of $\mathsf{PART}_\theta$. We propose to utilise a dictionary that stores the most commonly occurring subgraphs. Concretely, we define a dictionary $D$ to be a collection of connected subgraphs (or \textit{atoms}) from some universe $\univ$:
\begin{align}
    D = \{a_1, a_2, \cdots, a_{|D|}\}, \quad \text{where} \quad a_i \in \univ.
\end{align}

There are two viable choices for the atom encoding: 
(a) If the universe is small enough to be efficiently enumerable then we can assume a uniform distribution over $\univ$ which yields the description length $\length(D) = |D|\log|\univ|$. Intuitively, this would amount to storing the index of each atom within a list enumerating $\univ$. (b) On the other hand, when $\univ$ is too large to enumerate the atoms can be stored one-by-one based on the null-model encoding given in \eqref{eq:null}:
\begin{align}
    \length(D) = \sum_{a_i \in D} \length_{\text{null}}(a_i).
\end{align}
It is important to note that, to be as effective as possible, the partitioning and the dictionary should be co-designed: $D$ should capture those subgraphs that are most likely, whereas $\mathsf{PART}_\theta$ should be biased towards subgraphs with similar structure. 
The use of a dictionary makes it possible to explicitly account for (and thus optimise) the description  length of the learned hypothesis (i.e., equation above), which is an essential component of any compression algorithm (see challenge C3 in Section~\ref{sec:intro}).

\subsection{Step 3. Graph encoding}\label{sec:encoding}

The last step entails compressing $G$ by encoding the output $(\cH, C)$ of $\mathsf{PART}_\theta$.
As discussed in Section \ref{sec:prelims}, a uniquely decodable code implies a probability distribution, i.e., $q_\phi(G) =  q_\phi(\mathcal{H}, C)$ which corresponds to a description length ${\length_\phi(\mathcal{H}, C) = -\log q_\phi (\mathcal{H}, C})$, where $\phi$ denotes the learnable parameters.
In the following we explain how $q_\phi$ is decomposed.

\paragraph{Subgraphs.} We opt for a dual encoding of subgraphs: one for the subgraphs that belong to the dictionary $\cH_{\text{dict}}$, and one for non-dictionary subgraphs $\cH_{\text{null}}$ that are encoded with the help of a null model as in Eq. \eqref{eq:null}. This choice has a dual purpose: (a) It allows $\mathsf{PART}_\theta$ to choose non-dictionary atoms. This is crucial to our approach, since constraining the partitioning to specific isomorphism classes would significantly complicate optimisation. (b) Further, it enables us to maintain a balance between two common structures found in real-world networks---\textit{frequent subgraphs} (stored in the dictionary), and \textit{low-entropy subgraphs} as implied by the null model (i.e., very sparse or very dense subgraphs). The distribution is therefore decomposed into the following components:

\textit{Number of subgraphs. } First, we encode the number of dictionary and non-dictionary subgraphs ($b_{\text{dict}}$ and $b_{\text{null}}$ respectively) as follows:
\begin{equation}
    q_\phi(b_{\text{dict}}, b_{\text{null}}) 
    =  \text{Binomial}(b_{\text{dict}}|b; \phi) q_\phi(b) = {b \choose b_{\text{dict}}}(1-\delta_\phi)^{b_{\text{dict}}}\delta_\phi^{b-b_{\text{dict}}}q_\phi(b),
\end{equation}
where $1 - \delta_\phi = \mathbb{P}[H \in D]$ is the probability of an arbitrary subgraph to belong in the dictionary and $q_\phi(b)$ is a categorical.

\textit{Dictionary subgraphs. } The dictionary subgraphs are encoded in a permutation invariant way via a \textit{multinomial} distribution, i.e., we encode the histogram of atoms:
\begin{equation}
    q_\phi(\cH_{\text{dict}}|b_{\text{dict}}, D) = \text{Multinomial}(b_1, b_2, \dots, b_{|D|}\ |\ b_{\text{dict}};\phi) = b_{\text{dict}}!\prod_{a\in D}\frac{q_\phi(a)^{b_a}}{b_a!} ,
\end{equation}
where $b_a = \big|\{H \in \cH_{\text{dict}} | H \cong a\}\big|$ and $\sum_{a \in D} b_a = b_{\text{dict}}$.

\textit{Non-Dictionary subgraphs. } The non-dictionary subgraphs are encoded independently according to the null model:
\begin{equation}
    q_\phi(\cH_{\text{null}}|b_{\text{null}}, D) = \prod_{H_i \in \cH_{\text{null}}}q_{\text{null}}(H_i).
\end{equation}

\paragraph{Cuts.} We encode the cuts conditioned on the subgraphs, using a non-parametric uninformative null model for multi-partite graphs similar to \cite{peixoto2013parsimonious} (see Appendix \ref{sec:cut_encoding} for the detailed expression) %
 that prioritises low-entropy cuts.  In this way we give more emphasis to the subgraphs and an inductive bias towards distinct clusters in the graph. %

Overall, the description length of $(\cH, C) = \mathsf{PART}_\theta(G) $ is given by 
\begin{align}
    \hspace{-0mm}\length_\phi(\cH, C|D) 
    &= \length_\phi(b_{\text{dict}}, b_{\text{null}}) + \length_\phi(\cH_{\text{dict}}|b_{\text{dict}}, D) + \length_{\text{null}}(\cH_{\text{null}}|b_{\text{null}}, D) + \length_{\text{null}}(C|\cH), 
\end{align}
and the learnable parameter set is $\bigg\{\delta_\phi, \{q_\phi(b)\big\}_{b=b_{\text{min}}}^{b_{\text{max}}}, \{q_\phi(a)\}_{a \in D}\bigg\}$.

\paragraph{Remarks about graph isomorphism (GI). }Observe that given a fixed decomposition, our parametrisation is invariant to isomorphism, which is a desirable property since isomorphic graphs will be assigned codewords with the same length. Note that this is not sufficient to guarantee that all isomorphic graphs will be assigned \textit{the same} codeword. Since this would imply a solution to GI, it remains an open problem.  However, our graph encoding provides desirable tradeoffs between the expressivity of the probabilistic model, its number of parameters, and computational complexity, since it can adapt to different graph distributions using only a few parameters and solving GI only for small graphs.

\subsection{Selecting a hypothesis by minimising the total description length}
\label{sec:objective}

Putting everything together, in order to encode a graph dataset $\mathcal{G}$ sampled i.i.d. from $\mathfrak{G}$ we minimise the total description length %
\begin{align}\label{eq:MDL_general}
\min_{\theta, D, \phi} \ \sum_{G\in\mathcal{G}}\length_\phi(\mathsf{PART}_\theta(G)\, |\, D) + \length(D)
\end{align}
with respect to the parameters $\theta$ of the parametric partitioning algorithm, the dictionary $D$, and the parameters $\phi$ of the probabilistic model.
Eq. \eqref{eq:MDL_general} is a typical two-part Minimum Description Length (MDL) objective \cite{rissanen1978modeling, grunwald2007minimum}. Using standard MDL terminology, the tuple $(\theta, D, \phi)$ is a \textit{point hypotheses} and minimising \eqref{eq:MDL_general} amounts to finding the simplest hypothesis that best describes the data.

\section{Theoretical analysis: quadratic and linear gains}

The following section performs a comparative analysis of the description length growth rate of various graph compressors.
We compare PnC against two strong baselines: (a) The code length $\length_{\text{part}}$ induced by a pure partitioning-based graph encoding. Here, a graph is decomposed into subgraphs and cuts, but the distribution of subgraphs is not modelled (i.e., both subgraphs and cuts are encoded with a null model $\length_{\text{part}}(G) = \length_{\text{null}}(\cH) + \length_\text{null}(C|\cH)$). (b) The code length of encodings that do not rely on partitioning but encode each graph as a whole. Importantly, our results hold even for those baselines that encode the isomorphism class of each graph, rather than the graph itself, such as the Erd\H{o}s-Renyi model for unlabelled graphs of $n$ vertices:
$
\length_{\text{ER-S}}(G) = \log|\mathfrak{G}_{n,m}| + \log (n^2 + 1),  %
$
where $\mathfrak{G}_{n,m}$ is the set of all graphs with $n$ vertices and $m$ edges. $\length_{\text{ER-S}}(G)$ serves as lower bound to typical encodings such as that of Eq.~\eqref{eq:null}, but can be impractical to implement due to the complexity of GI. The analysis of additional baselines can be found in Appendix~\ref{sec:theory}.

Our main theorem shows that, under mild conditions on the underlying graph distribution, the expected description lengths of the compared encodings are totally ordered: %
\begin{theorem}\label{thrm:main}
Consider a distribution $p$ over graphs with $n$ vertices and a partitioning algorithm that decomposes a graph into $b$ blocks of $k = O(1)$ vertices. Then it holds that:
 \begin{align}
     \E_{G \sim p}[\length_{\textnormal{PnC}}(G)]
     \stackrel{(1b)}{\lesssim}
     \E_{G \sim p}[\length_{\textnormal{part}}(G)]
     \stackrel{(1a)}{\lesssim}
     \E_{G \sim p}[\length_{\textnormal{ER-S}}(G)]
 \end{align}
under the following conditions:
    
    (1a) $\frac{\log(k^2 + 1)}{k^2}) +\bar{\textnormal{\entropy}}_{m_{ij}} < \bar{\textnormal{\entropy}}_m $, where  \textnormal{$\bar{\entropy}_{m_{ij}} = \E_{G \sim p}[ \entropy\big(\frac{m_{ij}}{k^2}\big)]$} and \textnormal{$\bar{\entropy}_m = \E_{G \sim p}[\entropy(\frac{m}{n^2})]$} is the expected binary entropy of the cut size $m_{ij}$ between two subgraphs and that of the total number of edges $m$, respectively.

    (1b) $|D| <  (k^2 + 1)2^{k^2\bar{\textnormal{\entropy}}_{m_i} }$, where $|D|$ is the size of the dictionary and \textnormal{$\bar{\entropy}_{m_{i}} = \E_{G \sim p}[ \entropy\big(\frac{m_i}{k^2}\big)]$} the expected binary entropy of the number of edges $m_i$ in a subgraph.

The 
compression gains are:
  \begin{align}
    \E_{G \sim p}[\length_{\textnormal{Part}}(G)] 
     &\lesssim \E_{G \sim p}[\length_{\textnormal{ER-S}}(G)] - n^2\Big(\bar{\textnormal{\entropy}}_m - \frac{\log(k^2 + 1)}{k^2} -
     \bar{\textnormal{\entropy}}_{m_{ij}}\Big)\\
     &\quad\quad\quad\quad\quad\quad \textnormal{ and }\nonumber\\
    \E_{G \sim p}[\length_{\textnormal{PnC}}(G)] & \lesssim \E_{G \sim p}[\length_{\textnormal{part}}(G)]
    - nk(1-\delta)\bigg( \bar{\textnormal{\entropy}}_{m_i} - \frac{\mathbb{H}(D) - \log(k^2 + 1)}{k^2}\bigg),
 \end{align}
where $1 - \delta$ is the probablity that a subgraph belongs in the dictionary and $\mathbb{H}(D) = \mathbb{H}_{a\sim q_\phi(a)}[a]$ is the entropy of the distribution on dictionary atoms $q_{\phi}(a)$.
\end{theorem}
Theorem \ref{thrm:main} provides insights on the compressibility of certain graph distributions given their structural characteristics. In particular, we can make the following remarks:
(a) Condition (1a) can be satisfied even for very small values of $k$ as long as the graphs possess community structure. Perhaps counter-intuitively, when $k=O(1)$ we can satisfy the condition even if the communities have $O(n)$ size by splitting them into smaller subgraphs. This is possible because, in contrast to the majority of graph partitioning objectives that are based on minimum cuts, the compression objective attains its minimum when the cuts have ``low entropy''. Since communities that are tightly internally connected have large cuts, $\bar{\entropy}_{m_{ij}}$ and the code length will be kept small. This is a key observation that strongly motivates the use of partitioning for graph compression. 
(b) Condition (1b) provides an upper bound to the size of the dictionary,  which can be easily satisfied for moderately small values of $k$. More importantly, the dependence of the compression gain on the entropy $\mathbb{H}(D)$, reveals that dictionary atoms should be frequent subgraphs in the distribution, confirming our intuition. The bounds also show that, since the probabilities of the atoms are estimated from  the data, PnC does not need to make assumptions about the inner structure of the subgraphs and can adapt to general distributions. 

In Appendix \ref{sec:isomorphism_compression}, we also provide theoretical evidence on the importance of of encoding dictionary subgraphs as isomorphism classes instead of adjacency matrices, which is related to challenge C1 mentioned in the introduction. In particular, Theorem~\ref{thrm:isomorphism_compression} shows that, if isomorphism is not taken into account, the number of bits that will be lost will grow linearly with the number of vertices. All proofs and detailed assumptions can be found in Appendix \ref{sec:theory}.

\section{Optimisation and learning algorithms}

We turn our focus to learning algorithms for the optimisation of the MDL objective~\eqref{eq:MDL_general}. The following sections explain how each parametric component of PnC is learned.  %

\textbf{Subgraph encoding $\phi$.}
The graph encoding is parametrised as follows:  $q_\phi(a_i)$ and $q_\phi(b)$ are parametrised by learnable variables that are converted into categorical distributions over the dictionary atoms and the number of vertices respectively, using a softmax function. Similarly, $\delta_\phi$ is parametrised by a learnable variable converted to a probability via the sigmoid function. 

\textbf{Dictionary $D$.}
Let $\univ = \{a_1, a_2, \ldots a_{|\univ|}\}$ be a practically enumerable universe and define ${\bx = (x_1, x_2, \ldots, x_{|\univ|})}$ as
$$
    x_i = 
    \begin{cases} 
      1       & \text{if } a_i \in D \\
      0       & \text{otherwise}. 
  \end{cases}
$$  
Thus, $x_i$ indicates whether $D$ contains subgraph $a_i$. Now, optimising w.r.t the dictionary amounts to finding the binary assignments for $\bx$ that minimise~\eqref{eq:MDL_general}. To circumvent the combinatorial nature of this problem, we apply the continuous relaxation $\hat{x}_i = \sigma(\psi_i), \forall i \in \{0,1,\ldots, |\univ|\}$, where $\sigma$ is the sigmoid,
$\psi_i$ are learned continuous variables, and $\hat{x}_i \in [0,1]$  a fractional alternative to $x_i$. 
Appendix \ref{sec:relaxation} shows how~\eqref{eq:MDL_general} can be re-written w.r.t. $\bx$ and optimised by using the surrogate gradient w.r.t $\hat{x}_i$.

It is important to note that, in practice, we do not have to introduce indicator variables for the entire universe: Since most subgraphs $a_i$ will be never encountered in the graph distribution, we build the universe adaptively during training, by progressively adding the different graphs that the partitioning algorithm yields. We also allow the universe to contain subgraphs of size up to $k=O(1)$, in order to ensure that the isomorphism testing between atoms and subgraphs can be efficiently computed.

\textbf{Parametric graph partitioning algorithm.}
Finding the graph partitions that minimise~\eqref{eq:MDL_general} in principle requires searching in the space of 
partitioning algorithms. Instead, we constrain this space via a differentiable parametrisation that allows us to perform  gradient-based optimisation. 
Currently, learning to partition is an open problem, as to the extent of our knowledge known neural approaches require a fixed number of clusters \cite{DBLP:conf/nips/WilderEDT19, nazi2019gap, bianchi2019mincut} or do not guarantee that the subgraphs are connected~\cite{karalias2020erdos}.

Our \textit{Neural Partitioning} is a randomised algorithm 
parametrised with a graph neural network (GNN). When run on a graph, the GNN outputs a random $(\cH,C)$ together with a corresponding probability $p_\theta^{\text{GNN}}(\cH,C|G)$
and training is performed by estimating the gradients w.t.t. $\theta$ with REINFORCE \cite{williams1992simple}.
Our algorithm proceeds by iteratively sampling (and removing) subgraphs from the graph until it becomes empty.
At each step $t$ we select a subgraph $H_t$, by first sampling its vertex count $k_t$, and subsequently  sampling at most $k_t$ vertices.
To guarantee connectivity, we also sample the vertices iteratively and mask-out the probabilities outside the pre-selected vertices' neighbourhoods. The complexity of the algorithm is $O(n)$, where $n$ the number of the vertices of the graph. 
Please refer to Appendix \ref{sec:neural_part} for an in-depth explanation of the algorithm and  relevant implementation details. We stress that we mainly consider this algorithm as a proof of concept that we ablate against other non-parametric partitioning algorithms. A plethora of  solutions can be explored in a parametric setting and we welcome future work in this direction.

\textbf{MDL objective.} Given dataset $\cG$, we train all components by minimising the description length:
\begin{align}\label{eq:MDL_general_param}
L(\cG) = \length_{\bx}(D) + \sum_{G\in\mathcal{G}}\E_{(\cH,C)\sim p^{\text{GNN}}_\theta(\cH, C|G)}[\length_{\phi,\bx}(\cH,C|D)].
\end{align}
Taking the expectation over the GNN output $(\cH,C)\sim p^{\text{GNN}}_\theta(\cH, C|G)$), we calculate
the gradients as:
${\nabla_\phi L(\cG) = \sum_{G\in\mathcal{G}}\E[\nabla_\phi\length_{\phi,\bx}(\cH,C)|D)]}$, 
${\nabla_{\hat{\bx}} L(\cG) = \nabla_{\hat{\bx}}\length_{\hat{\bx}}(D) + \sum_{G\in\mathcal{G}}\E[\nabla_{\hat{\bx}}\length_{\phi,\hat{\bx}}(\cH,C|D)]}$,
 and
${\nabla_\theta L(\cG) = \sum_{G\in\mathcal{G}}\E[\length_{\phi,\bx}(\cH,C|D)\nabla_\theta\ln p^{\text{GNN}}_\theta(\cH,C|G)]}$.

\begin{figure}[t]
    \begin{minipage}[t]{0.48\linewidth}
        \centering
          \includegraphics[width=\linewidth]{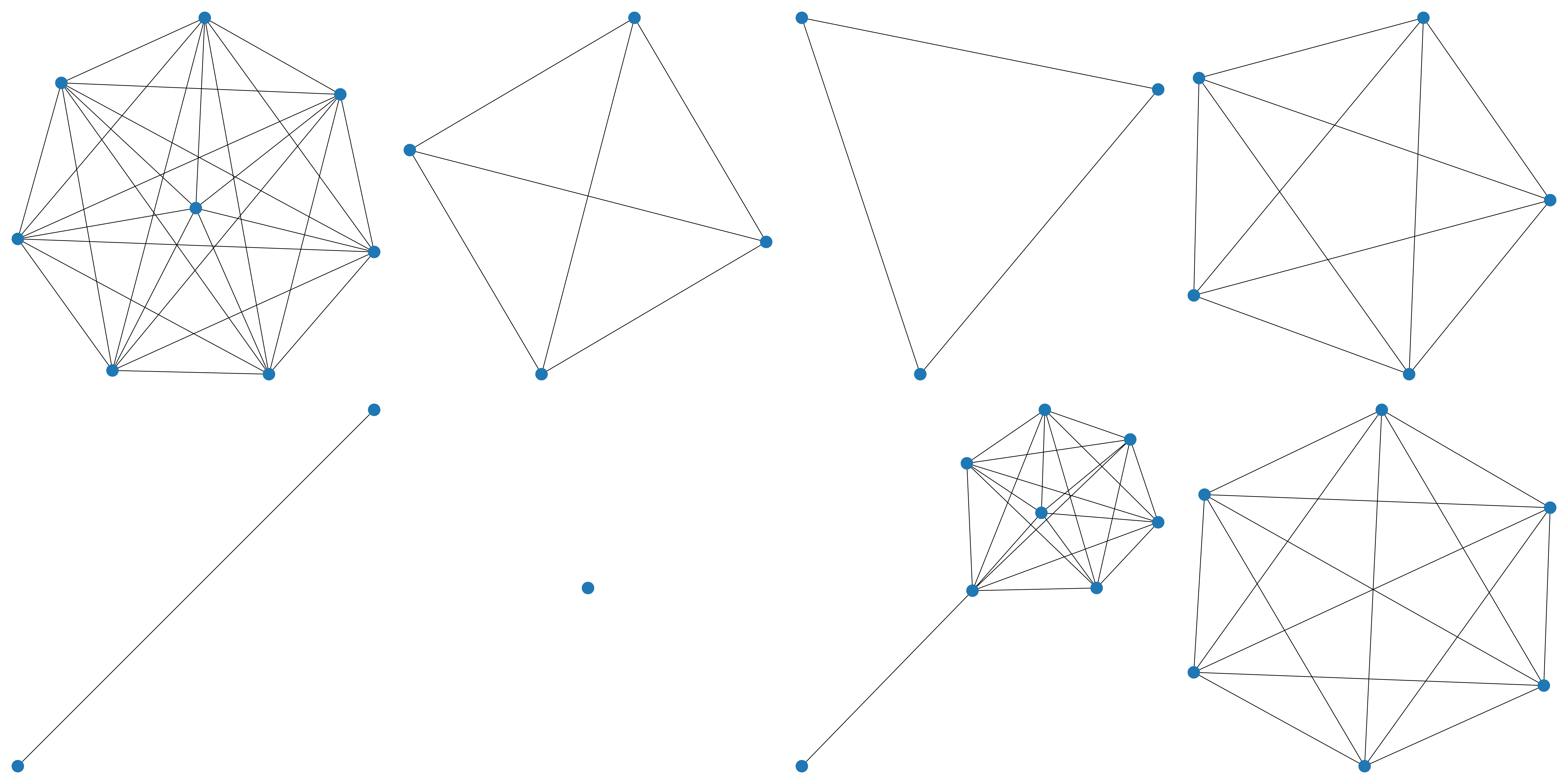}
        \label{fig:zinc-substructures}
    \end{minipage}
    \hfill
    \begin{minipage}[t]{0.48\linewidth}
        \centering
         \includegraphics[width=\linewidth]{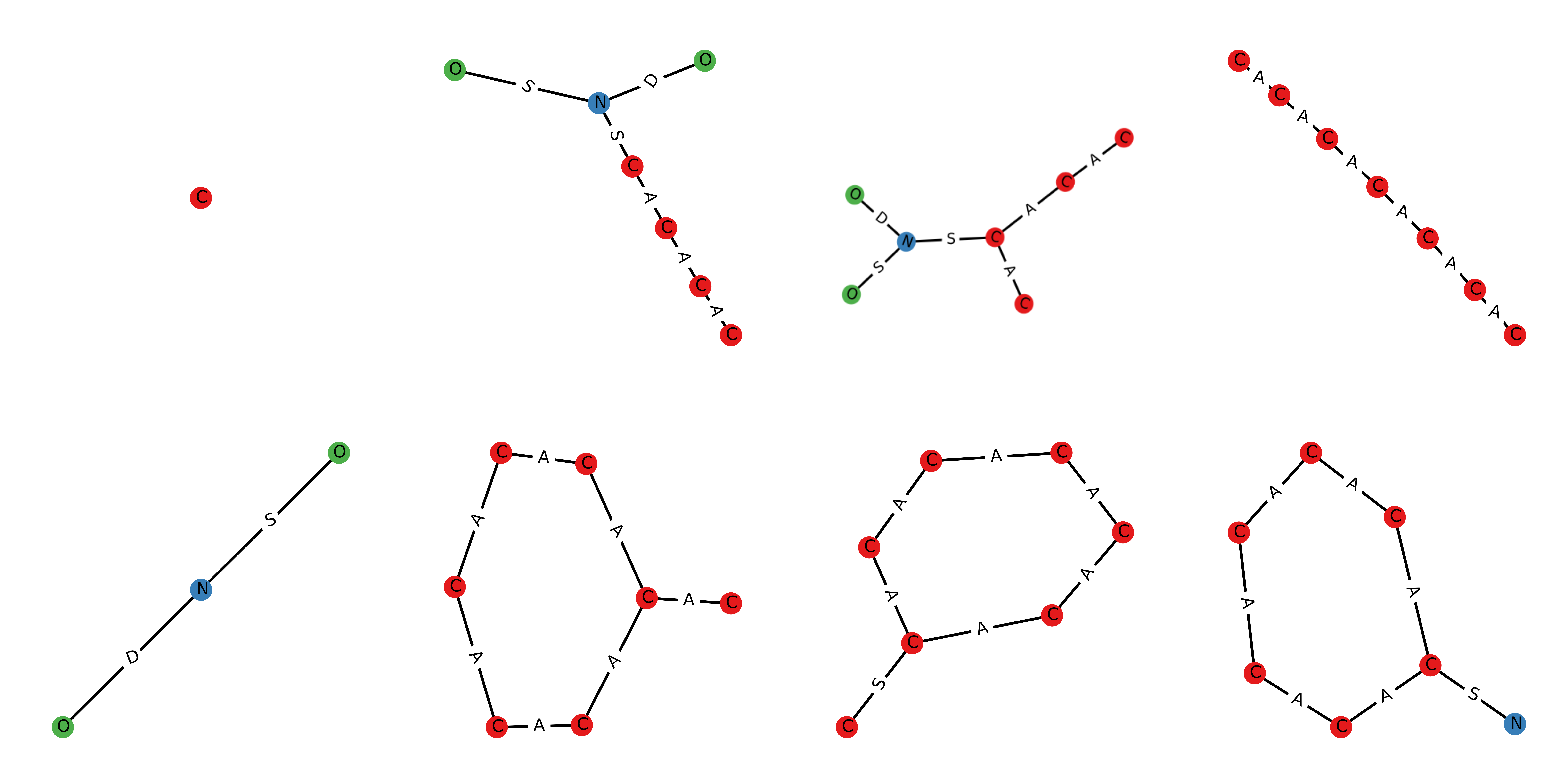}
        \label{fig:sr_plot} 
    \end{minipage}
    \caption{PnC + Neural Part. -  Most probable graphs in the IMDB-B dataset (left) and the attributed MUTAG dataset (right). Atom types and bond types
    are represented as vertex and edge attributes.
    }
    \label{fig:substructures.}
\end{figure}

\section{Empirical Results}\label{sec:main_evaluation}

We evaluate our framework in a variety of datasets: small molecules, proteins and social networks~\cite{yanardag2015deep,irwin2012zinc,Morris+2020,borgwardt2005protein,helma2001predictive}. Across all methods, we assume an optimal encoder that attains the entropy lower bound (this is often realistic since modern entropy coders asymptotically approach it)
to evaluate the expressive power of each model independently of the encoder.
We measure the description length of the data as their negative log-likelihood (NLL) under each probabilistic model, as well as the total description length by adding the cost of the parameters which need to be transmitted to the decoder (see Appendix \ref{sec:implementation} for details).

\textbf{Baselines.} We aim to assess representative approaches across the entire spectrum of graph probabilistic models, i.e., from completely uninformative non-parametric distributions to overparametrised neural generative models. We consider the following types of compressors:
(a) \textit{Null models.} 
We select the \textit{uniform} model, where all edges are assumed to be sampled independently with probability equal to 0.5, the \textit{edge list} model, a typical graph representation,  and the \textit{Erd\H{o}s-Renyi} model (Eq. \eqref{eq:null}). %
(b) \textit{Partitioning-based.} non-parametric methods that aim at grouping vertices in tightly-connected clusters. They can be used for any type of sparse matrix \cite{chakrabarti2004fully} and are based on the assumption that there exists a hidden community structure in the graph. %
The partioning algorithms used are \textit{SBM fitting} \cite{peixoto2012entropy,peixoto2013parsimonious, peixoto2017nonparametric, peixoto2019bayesian},  Louvain  \cite{blondel2008fast} and Label Propagation \cite{raghavan2007near} clustering. The encoding we use to encode the clusters corresponds exactly to the SBM assumptions, hence the partitioning-based results are always superior for this approach. 
(c) \textit{Likelihood-based neural compressors.} As with any likelihood-based model, graph generative models can be transformed into graph compressors. We evaluate the original GraphRNN \cite{you2018graphrnn} and GRAN \cite{liao2019efficient} networks, as well as smaller instantations that have undergone model compression using the Lottery Ticket Hypothesis algorithm \cite{DBLP:conf/iclr/FrankleC19}.

\textbf{Results.}  Tables \ref{tab:mol} and \ref{tab:tab-social} report the compression quality of each method measured in terms of the average number of bits required to store each edge in a dataset (bpe). We present four variants of PnC, differing on the type of partitioning algorithm used~\cite{raghavan2007near,blondel2008fast,peixoto2019bayesian}. 
We report separately the cost of compressing the data as well as the total cost (including the  parameters). %
Several observations can be made with regards to the baselines: 

First off, \textit{off-the-shelf likelihood-based neural approaches are poor compressors due to failing to address challenge C3.}  These models exhibit an unfavorable trade-off between the data and model complexity, often requiring significantly more bpe than the null models. Although model compression techniques can alleviate this tradeoff (especially for larger datasets, e.g., pruned GraphRNN on ZINC), in most of the cases the compression ratios required to outperform PnC are significantly higher than the best that have been reported in the literature (See Table \ref{tab:ratios} in the Appendix). Perhaps more importantly,
it is unclear how to optimise the model description length during training (one of the few exceptions is \cite{DBLP:conf/iclr/HavasiPH19}) and usually model compression might be tedious and is based on heuristics
\cite{DBLP:journals/corr/HanMD15, DBLP:conf/nips/LouizosUW17, DBLP:conf/nips/HanPTD15, hinton2015distilling}. See Appendix \ref{sec:ablation_likelihood} for more details and additional experiments.

In addition, as expected from Theorem (1a), \textit{non-parametric clustering algorithms work well when the dataset has a strong community structure, but are not a good choice for more structured datasets.} For instance, the best clustering algorithm requires 2.4$\times$ more bpe for ZINC than the best PnC.  
\textit{PnC variants achieve the best compression in all datasets considered.} This follows from Theorem (1b), since the learned dictionaries are relatively small, and confirms our hypothesis that our framework is sufficiently flexible to account for the particularities of each dataset. As seen, neural partitioning performs in every case better, or on par with the combination of PnC with the Louvain algorithm. However, in the social network datasets, the combination of PnC with SBM achieves the best performance. This occurs because these networks fit nicely with the SBM inductive bias (which as a matter of fact is exactly that of low-entropy cuts), and most importantly, due to the fact that the clusters recovered by the SBM are small and repetitive,
which makes them ideal for the PnC framework. We also observe that there is room for improvement for the neural partitioning variant, and hypothesise that a more powerful parametrised algorithm can be designed. Since learnable partitioning is still an open problem, we leave this research direction to future work.

Fig. \ref{fig:substructures.} shows the most likely dictionary atoms for the IMDB-B and the MUTAG dataset (also including attributes -  Appendix \ref{sec:attrs} provides additional experiments). Observe that cliques or near-cliques and typical molecular substructures, such as carbon cycles and junctions are recovered for social networks and molecules respectively. This clearly highlights the connection between compression and pattern mining and provides evidence for potential applications of our framework.

\begin{table}[t]
\caption{Average bits per edge (bpe) for molecular graph datasets. \textcolor{red}{\textbf{First}}, \textcolor{violet}{\textbf{Second}}, \textcolor{blue}{\textbf{Third}}}
\label{tab:mol}
\resizebox{\textwidth}{!}{%
\begin{tabular}{@{}llccccccccc@{}}
\toprule
\multirow{3}{*}{\begin{tabular}[c]{@{}l@{}}Method\\ type\end{tabular}} &
  Graph type &
  \multicolumn{9}{c}{Small Molecules} \\ \cmidrule(l){2-11} 
                      & Dataset name            & \multicolumn{3}{c}{MUTAG}        & \multicolumn{3}{c}{PTC}       & \multicolumn{3}{c}{ZINC}          \\ \cmidrule(l){2-11} 
                      &                         & data & total            & params & data       & total   & params & data & total             & params \\ \midrule
\multirow{3}{*}{Null} & Uniform (raw adjac.) & -    & 8.44             & -      & - & 17.43   & -      & -    & 10.90             & -      \\
                      & Edge list               & -    & 7.97             & -      & -          & 9.38    & -      & -    & 8.60              & -      \\
                      & Erd\H{o}s-Renyi             & -    & 4.78             & -      & -          & 5.67    & -      & -    & 5.15              & -      \\ \midrule
Partitioning            & SBM-Bayes               & -    & 4.62            & -      & -          & 5.12    & -      & -    & 4.75              & -      \\
(non-parametric)      
& Louvain                 & -    & 4.80             & -      & -          & 5.27    & -      & -    & 4.77              & -      \\
& PropClust               & -    & 4.92             & -     & -          & 5.40    & -       & -    & 4.85              & -       \\ \midrule
Neural                & GraphRNN                
& 1.33 & 3338.21           & 388K 
& 1.57       & 1394.59  & 389K   
& 1.62 & 43,16          & 388K   \\
(likelihood)          & GRAN                   
&  0.81 &   12557.75                   &  1460K  
&  2.18 &  5269.82        &    1470K
& 1.30     & 157.7                 &      1461K  \\
   & GraphRNN (pruned) & 1.95 & 12.39 & 1.08K & 2.16 & 6.71 & 1.10K & 1.79 & { \color{violet}\textbf{2.02}} & 1.90K\\
   & GRAN (pruned) & 2.59  & 24.56 & 2.23K & 4.31 &14.00 &2.36K &3.26 & 3.47 & 1.69K \\
\midrule
PnC                    
& PnC + SBM               & 3.81   & 4.11    & 49    & 4.38       & 4.79  & 155 & 3.34   & 3.45    & 594   \\
 &
  PnC + Louvain &
  2.20 &
  {\color{violet} \textbf{2.51}} &
  47 &
  2.68 &
  {\color{violet} \textbf{3.14}} &
  166 &
  1.96 &
  {\color{red} \textbf{1.99}} &
  196 \\
 &
  PnC + PropClust &
  2.42 &
  {\color{blue} \textbf{3.03}} &
  63 &
  3.38 &
  {\color{blue} \textbf{4.02}} &
  178 &
  2.20 &
  {2.35} &
  726 \\ \midrule
 &
  PnC + Neural Part. &
  2.17±0.02 &
  {\color{red} \textbf{2.45±0.02}} &
  46±1 &
  2.63±0.26 &
  {\color{red} \textbf{2.97±0.14}} &
  143±31 &
  2.01±0.02 &
  {\color{blue} \textbf{2.07±0.03}}&
  384±105 \\ \bottomrule%
\end{tabular}%
}
\end{table}

\begin{table}[t]
\caption{Average bits per edge (bpe) for social and protein graph datasets. \textcolor{red}{\textbf{First}}, \textcolor{violet}{\textbf{Second}}, \textcolor{blue}{\textbf{Third}}}
\label{tab:tab-social}
\resizebox{\textwidth}{!}{%
\begin{tabular}{@{}llccccccccc@{}}
\toprule
\multirow{3}{*}{\begin{tabular}[c]{@{}l@{}}Method\\ type\end{tabular}} &
  Graph type &
  \multicolumn{3}{c}{Biology} &
  \multicolumn{6}{c}{Social Networks} \\ \cmidrule(l){2-11} 
 &
  Dataset name &
  \multicolumn{3}{c}{PROTEINS} &
  \multicolumn{3}{c}{IMDB-B} &
  \multicolumn{3}{c}{IMDB-M} \\ \cmidrule(l){2-11} 
   &
   &
  data &
  total &
  params &
  data &
  total &
  params &
  data &
  total &
  params \\ \midrule
\multirow{3}{*}{\begin{tabular}[c]{@{}l@{}}Null\end{tabular}} &
  Uniform (raw adjac.) &
  - &
  24.71 &
  - &
  - &
  2.52 &
  - &
  - &
  1.83 &
  - \\
 &
  Edge list &
  - &
  10.92 &
   &
  - &
  8.29 &
  - &
  - &
  7.74 &
  - \\
 &
  Erd\H{o}s-Renyi &
  - &
  5.46 &
  - &
  - &
  1.94 &
  - &
  - &
  1.32 &
  - \\ \midrule
\multirow{3}{*}{\begin{tabular}[c]{@{}l@{}}Partitioning\\ (non-parametric)\end{tabular}} &
  SBM-Bayes &
  - &
  3.98 &
  - &
  - &
  {\color{violet} \textbf{0.80}} &
  - &
  - &
 {\color{violet} \textbf{ 0.60}} &
  - \\
 &
  Louvain &
  - &
  3.95 &
  - &
  - &
  1.22 &
  - &
  - &
  0.88 &
  - \\
 &
  PropClust &
  - &
  4.11 &
  - &
  - &
  1.99 &
  - &
  - &
  1.37 &
  - \\ \midrule
\multirow{2}{*}{\begin{tabular}[c]{@{}l@{}}Neural\\ (likelihood)\end{tabular}} &
  GraphRNN &
  2.03 &
   156.99  &
  392K &
  1.03 &
  132.27 &
  395K &
  0.72 &
   127.84 &
  392K \\
 &
  GRAN &
  1.51 &
   607.96  &
  1545K &
  0.26 &
  488.88 &
  1473K &
  0.17 &
  475.13 &
  1467K \\
     & GraphRNN (pruned) & 2.63 & 3.76 & 2.56K &1.43 & 1.92& 1.28K & 0.91 & 1.39 & 1.28k\\
   & GRAN (pruned) & 4.28 & 5.11 & 1.78K & 0.84 & 1.75 & 2.38K & 0.55 & 1.41 &2.31K \\
  \midrule
PnC &
  PnC + SBM &
  3.26 &
  {\color{blue} \textbf{3.60}} &
  896 &
  0.50 &
  {\color{red} \textbf{0.54}} &
  198 &
  0.38 &
  {\color{red} \textbf{0.38}} &
  157 \\
 &
  PnC + Louvain &
  3.34 &
  {\color{violet} \textbf{3.58}} &
  854 &
  0.96 &
  {\color{blue} \textbf{1.02}} &
  202 &
  0.66 &
  {\color{blue} \textbf{0.70}} &
  141 \\
 &
  PnC + PropClust &
  3.42 &
  {3.68} &
  866 &
  1.45 &
  1.64 &
  241 &
  0.93 &
  1.04 &
  178 \\ \cmidrule(l){2-11} 
 &
  PnC + Neural Part. &
  3.34±0.25 &
  {\color{red} \textbf{3.51±0.23}} &
  717±61 &
  1.00±0.04 &
  1.05±0.04 &
  186±25 &
  0.66±0.05 &
  0.72±0.05 &
  178±14
  \\ \bottomrule
\end{tabular}%
}
\end{table}

\section{Conclusion}

This paper marks an important step towards learnable entropy-based graph compression. To the best of our knowledge, our work represents the first attempt to address the basic principles of parametric compressors of unlabelled graphs learned from observations. In addition, we suggest practical instantiations of our framework that can be trained with gradient-based optimisation, accounting for the total description length, hence aiming for the largest possible parsimony. A number of new research questions arise, such as how to design more expressive, albeit parsimonious
distribution estimators, how to improve the neural partitioning algorithms, and how to ensure scalability to single large networks that pose a significant challenge w.r.t the memory constraints of GPUs.  %
We hope that our work will inspire further research in this emerging research area.

\section*{Acknowledgements and Disclosure of Funding}
We would like to thank the anonymous reviewers for their valuable feedback and suggestions to improve our paper. This project was partially funded by the ERC Consolidator Grant No. 724228 - LEMAN. Giorgos Bouritsas is partially supported by a PhD scholarship from the Dpt. of Computing, Imperial College London.  Andreas Loukas and Nikolaos Karalias thank the Swiss National Science Foundation for supporting them in the context of the project “Deep Learning for Graph Structured Data”, grant number PZ00P2 179981. Michael Bronstein acknowledges support from Google Faculty awards and the Royal Society Wolfson Research Merit award.

{
\bibliographystyle{unsrt}
\bibliography{references}
}

\newpage

\appendix
\section{Theoretical analysis}\label{sec:theory}

\subsection{Preliminaries}

Recall the definition of the binary entropy ${\entropy(p) = -p\log p - (1-p)\log(1-p)}$. A useful approximation that will be used in the analysis is the following:
\begin{equation}
    \log{n \choose m} \approx n \, \entropy\bigg(\frac{m}{n}\bigg),
\end{equation}
which holds when both $n$ and $m$ grow at the same rate, i.e., $m\lesssim n$, and can be derived using Stirling's approximation. 

For the following comparisons, we will be considering a graph distribution with $n$ vertices. %
For better exposition, the analysis will be performed for directed graphs. The same trends in the bounds hold also for undirected graphs.

\paragraph{Conventional graph encodings.}
We consider two types of baseline graph encodings:
\begin{itemize}
    \item Uniform: $\length_{\text{unif-G}}(G) = n^2$. Here no assumptions are made about the graph; all graphs are considered to be equally probable.
    \item Erd\H{o}s-Renyi: ${\length_{\text{ER-G}}(G) =  \log {n^2 \choose m}} + \log (n^2 + 1) \approx n^2 \entropy_m + \log (n^2 + 1)$, where $m$ the number of edges and ${\entropy_m = \entropy \big(\frac{m}{n^2}\big)}$. This baseline is efficient at encoding graphs that are either very sparse or very dense.
\end{itemize}

Though the above encodings assign the same probability to isomorphic graphs, they map them to different codewords. Hence, they are redundant when dealing with unlabelled graphs. The following variants are more efficient by taking into account isomorphism:
\begin{itemize}
    \item Uniform - Isomorphism classes: $\length_{\text{unif-S}}(G) = \log|\mathfrak{G}_{n}| 
    \approx n^2 - n\log n$. 
    \item Erd\H{o}s-Renyi - Isomorphism classes: $\length_{\text{ER-S}}(G) = \log|\mathfrak{G}_{n,m}| + \log (n^2 + 1)  \approx  n^2 \entropy_m + \log (n^2 + 1)  - n \log n$,
\end{itemize}
where $\mathfrak{G}_{n}$ and $\mathfrak{G}_{n,m}$ are the set of all graphs with $n$ vertices and the set of all graphs with n vertices and $m$ edges, respectively. In both cases, we used the fact that asymptotically almost all graphs are rigid i.e., that they have only the trivial automorphism~ \cite{erdHos1963asymmetric}.

Observing that all four encodings asymptotically grow quadratically with the number of nodes
we can derive the following lemma:

\begin{lemma} Consider a graph distribution $p$ over graphs with $n$ vertices and denote by \textnormal{$\bar{\entropy}_m = \E_{G\sim p}[\entropy(\frac{m}{n^2})]$} the expected value of the binary entropy of the number of edges $m$. If \textnormal{$\bar{\entropy}_m < 1$}, then the expected description lengths of the baseline models are asymptotically ordered as follows:
 \begin{align}
     \E_{G \sim p}[\length_{\textnormal{ER-S}}(G)] 
    \lesssim  \E_{G \sim p}[\length_{\textnormal{ER-G}}(G)]
     \lesssim \E_{G \sim p}[\length_{\textnormal{unif-S}}(G)] 
     \lesssim \E_{G \sim p}[\length_{\textnormal{unif-G}}(G)]
 \end{align}
The compression gain when encoding isomorphism classes instead of labelled graphs is $\Theta(n\log n)$, while that of the Erd\H{o}s-Renyi encoding compared to the uniform one is \textnormal{$\Theta\big(n^2(1-\bar{\entropy}_m)\big)$}.
\end{lemma}

The proof follows directly from the equations above. Since most real-world graphs are sparse, then the condition $\bar{\entropy}_m < 1$ is almost always true.

\paragraph{Partitioning (only).} The partitioning models considered in this analysis assume that each graph is clustered into $b$ subgraphs of $k$ vertices each (i.e., $b=\frac{n}{k}$) and that the intra- and inter-subgraph edges are encoded independently\footnote{To make the analysis more tanglible, we examine here a slightly simpler encoding than the one used in the experiments - see Eq. \eqref{eq:partitioning}.}. Note that the preamble terms that encode the number of blocks in the partition and the number of vertices per block are unnecessary since $k$ is fixed. Overall, the code length is given by 
$$
\length_{\text{Part}}(G) = \length(\cH) +  \length(C|\cH),
$$ 
with the two terms defined respectively as follows:
\begin{align}\label{eq:partitioning_simple}
\begin{split}
\length(\cH) &= \sum_{i=1}^b \bigg(\log (k^2 + 1) + \log {k^2 \choose m_{i}} \bigg)  
\quad \text{and} \quad 
\length(C|\cH) = \sum_{i\neq j}^b \bigg(\log (k^2 + 1) + \log {k^2 \choose m_{ij}}\bigg).   
\end{split}
\end{align}
Above, $m_i$ is the number of edges in subgraph $H_i$ and $m_{ij}$ is the size of the cut between subgraphs $H_i$ and $H_j$. In both cases, the first term encodes the number of edges, and the second their arrangement across the vertices.

\paragraph{Partition and Code (PnC).} We make the same assumptions for PnC as in $\length_{\text{Part}}(G)$: the graph is partitioned into $b$ subgraphs of $k$ vertices and the same encoding of non-dictionary subgraphs $\cH_{\null}$ and cuts $C$ is used.  The number of dictionary subgraphs $b_{\text{dict}}$ is encoded with a Binomial distribution and the dictionary subgraphs $\cH_{\text{dict}}$ themselves with a Multinomial as in Eq. (6) and (7) of the main paper. The overall encoding length is
\begin{align}\label{eq:pnc_simple}
\length_{\text{PnC}}(G|D) &= \length(b_{\text{dict}}) +  \length(\cH_{\text{dict}}|b_{\text{dict}}, D) + \length(\cH_{\text{null}}|b_{\text{null}}, D) +
\length(C|\cH),
\end{align}

Each dictionary atom is encoded using any of the null models mentioned above, hence $\length(D) = O(|D| k^2)$.

\subsection{Why partitioning? Partitioning vs Null Models}

As a warm-up, we will discuss the case of encodings based on pure graph partitioning, such as the Stochasic Block Model (``Partioning non-parametric'' in the tables of the main paper). We remind the reader that these encodings do \textit{not} take into account the isomorphism class of the identified subgraphs but rely on a null model to encode them. 

In the following, we derive a sufficient condition for the sparsity of the connections between subgraphs, under which partitioning-based encodings will yield smaller expected description length than the baseline null models. Formally:
\begin{customthm}{1a}\label{thrm:part_vs_null}
Let every $G \sim p$ be partitioned into $b$ blocks of $k = O(1)$ vertices and suppose that the partitioning-based encoding of Eq. \eqref{eq:partitioning_simple} is utilised. The following holds:
 \begin{align}
     \E_{G \sim p}[\length_{\textnormal{Part}}(G)] 
     \lesssim \E_{G \sim p}[\length_{\textnormal{ER-S}}(G)] - n^2\Big(\bar{\textnormal{\entropy}}_m - \frac{\log(k^2 + 1)}{k^2} -
     \bar{\textnormal{\entropy}}_{m_{ij}}\Big),
 \end{align}
where \textnormal{$\bar{\entropy}_{m_{ij}} = \E_{G\sim p}[ \entropy\big(\frac{m_{ij}}{k^2}\big)]$} and \textnormal{$\bar{\entropy}_m = \E_{G\sim p}[\entropy(\frac{m}{n^2})]$} are the expected binary entropy of the cuts and of the total number of edges, respectively.
\end{customthm}
\begin{proof}
\begin{align*}
    \E_{G\sim p}[\length_{\text{part}}(G)] &\approx \E_{G\sim p}\big[\sum_{i=1}^b \bigg(\log(k^2 + 1) + k^2\entropy\big(\frac{m_i}{k^2}\big)\bigg)
    + \sum_{i\neq j}^b \bigg(\log(k^2 + 1) + k^2\entropy\big(\frac{m_{ij}}{k^2}\big)\bigg)\big]\\
    &= \frac{n}{k} \bigg(\log(k^2 + 1) + k^2\bar{\entropy}_{m_i}\bigg) + (\frac{n^2}{k^2} - \frac{n}{k})\bigg(\log(k^2 + 1) + k^2\bar{\entropy}_{m_{ij}}\bigg)\\
    &=  n^2\bigg(\frac{\log(k^2 + 1)}{k^2} + \bar{\entropy}_{m_{ij}}\bigg) 
    + n k\bigg(\bar{\entropy}_{m_i}  - \bar{\entropy}_{m_{ij}}\bigg)\\
    &=\E_{G \sim  p}[\length_{\text{ER-S}}(G)] -
     n^2\bigg(\bar{\entropy}_m - \frac{\log(k^2 + 1)}{k^2} - \bar{\entropy}_{m_{ij}}\bigg)\\
     &+ n k\bigg(\bar{\entropy}_{m_i}  - \bar{\entropy}_{m_{ij}} + \log n \bigg) - \log (n^2 + 1)\\
    &\lesssim\E_{G \sim  p}[\length_{\text{ER-S}}(G)] -
     n^2\bigg(\bar{\entropy}_m - \frac{\log(k^2 + 1)}{k^2} - \bar{\entropy}_{m_{ij}} \bigg),
\end{align*}
where we assumed that $m \lesssim n^2$, $m_i \lesssim k^2$, $m_{ij} \lesssim k^2$, and in the last step we derive an asymptotic inequality using the dominating quadratic term.
In other words, 
partitioning-based encoding is quadratically superior to the best null model whenever there exists a $k$ such that 
$$
    \frac{\log(k^2 + 1)}{k^2} < \bar{\entropy}_m - \bar{\entropy}_{m_{ij}}.
$$
The above concludes the proof.
\end{proof}

\subsection{The importance of the dictionary: PnC vs Partitioning}

We proceed to mathematically justify why encoding subgraphs with a dictionary (the ``Code'' part in PnC) can yield extra compression gains compared to pure partitioning-based encodings. 

As our main theorem shows, utilising a dictionary allows us to reduce the linear $O(n)$ terms of the partitioning-based description length:
\begin{customthm}{1b}\label{thrm:pnc_vs_part}
Let every $G \sim p$ 
be partitioned into $b$ blocks of $k = O(1)$ vertices and suppose that the PnC encoding of Eq. \eqref{eq:pnc_simple} is used. If there exists a dictionary such that $|D| <  (k^2 + 1)2^{k^2\bar{\entropy}_{m_i} }$ with $\bar{\entropy}_{m_{i}} = \E_{G\sim p}[ \entropy\big(\frac{m_i}{k^2}\big)]$ being the expected binary entropy of the subgraph edges, then it holds that:
 \begin{align}
     \E_{G\sim p}[\length_{\textnormal{PnC}}(G)]  \lesssim \E_{G\sim p}[\length_{\textnormal{part}}(G)]
    - nk(1-\delta)\bigg(\bar{\textnormal{\entropy}}_{m_i} - \frac{\mathbb{H}(D) - \log(k^2 + 1)}{k^2}\bigg),
 \end{align}
where $1 - \delta$ is the probability that a subgraph belongs in the dictionary and $\mathbb{H}(D)$ is the entropy of the distribution $q$ over the dictionary atoms.
\end{customthm}

\begin{proof}
We will analyse the description length of each of the components of Eq. \eqref{eq:pnc_simple}.

\textit{Number of dictionary subgraphs.} The expected description length of the number of subgraphs is equal to the entropy of the binomial distribution:
\begin{align*}
    \E_{G\sim p}[\length(b_{\text{dict}})] 
    &= \frac{1}{2}\log\Big(2\pi e b \delta (1-\delta)\Big) + O\Big(\frac{1}{b}\Big) \\ 
    &= \frac{1}{2}\log\Big(2\pi e \delta (1-\delta)\Big) + \frac{1}{2}\log\Big(\frac{n}{k}\Big) + O\Big(\frac{k}{n}\Big) 
    = O\Big(\log \frac{n}{k}\Big)
\end{align*}

\textit{Dictionary subgraphs.} The expected description length of the subgraphs that belong in the dictionary amounts to the entropy of the multinomial distribution and can be upper bounded as follows: 
\begin{align*}
    \E_{G\sim p}[\length(\cH_{\text{dict}}|b_{\text{dict}}, D)] 
    &= \E_{G\sim p}\bigg[-\log \frac{b_{\text{dict}}!}{\prod_{a\in D} b_a!} - \sum_{a\in D}b_a\log q(a)\bigg] \\
    &\leq \E_{G\sim p}[-\sum_{a\in D}b_a\log q(a)]\\
    &=  -\sum_{a\in D}\E_{G\sim p}[b_a]\log q(a) \\
    &=   - b(1-\delta)\sum_{a\in D}q(a)\log q(a) 
    = \frac{n}{k}(1-\delta) \mathbb{H}(D) 
    \leq \frac{n}{k}(1-\delta)\log |D|,
\end{align*}
where we used the fact that $b_{\text{dict}}!\geq \prod_{a\in D} b_a!$.  The term ${\mathbb{H}(D) = \mathbb{H}_{a \sim q(a)}[a] = - \sum_{a\in D}q(a)\log q(a)}$ corresponds to the entropy of the dictionary distribution.

\textit{Non-dictionary subgraphs.} Assuming that the subgraph edges $m_i$ are independent from the number of non-dictionary subgraphs $b-b_{\text{dict}}$, then the expected description length of the non-dictionary subgraphs becomes:
\begin{align*}
    \E_{G\sim p}[\length(\cH_{\text{null}}|b_{\text{null}}, D)] \approx \E_{G\sim p}\big[\sum_{i=1}^{b - b_{\text{null}}} \bigg(\log(k^2 + 1) + k^2\entropy_{m_i}\bigg)] =  \frac{n}{k}  \delta \bigg(\log(k^2 + 1) + k^2\bar{\entropy}_{m_i}\bigg)
\end{align*}

Overall, using the same derivation for the cuts as in Theorem~\ref{thrm:part_vs_null}, we obtain:
\begin{align*}
\E_{G\sim p}[\length_{\text{PnC}}(G|D)] &\lessapprox  O(\log \frac{n}{k}) + n^2\bigg(\frac{\log(k^2 + 1)}{k^2} + \bar{\entropy}_{m_{ij}}\bigg) \\
    &+ n \bigg(\delta k \bar{\entropy}_{m_i} + (1-\delta)\bigg(\frac{\mathbb{H}(D) - \log(k^2 + 1)}{k}\bigg) - k\bar{\entropy}_{m_{ij}}\bigg)\\
    &= \E_{G\sim p}[\length_{\text{part}}(G)] 
    + n (1-\delta)\bigg(\frac{\mathbb{H}(D) - \log(k^2 + 1)}{k} - k \bar{\entropy}_{m_i}\bigg) + O(\log \frac{n}{k})
\end{align*}
Including the description length of the dictionary and amortising it over each graph in a dataset of $\cG$ graphs, we conclude
\begin{align*}
\E_{G\sim p}[\length_{\text{PnC}}(G)]
    &= \E_{G\sim p}[\length_{\text{part}}(G)] 
    - nk (1-\delta)\bigg(\bar{\entropy}_{m_i} - \frac{\mathbb{H}(D) - \log(k^2 + 1)}{k^2}\bigg) + O(\log \frac{n}{k} + \frac{|D|}{|\cG|}k^2).
\end{align*}
Hence, if $k = O(1)$ and $|D| \ll |\cG|$ (more precisely, the ratio $\frac{|D|}{|\cG|}$ shouldn't grow with $n$), then a linear compression gain is obtained if: 
\begin{align*}
    \bar{\entropy}_{m_i} - \frac{\mathbb{H}(D) - \log(k^2 + 1)}{k^2}> 0 \Leftrightarrow  \mathbb{H}(D) <  \log (k^2 + 1) + k^2\bar{\entropy}_{m_i}. 
\end{align*}
The proof concludes by noting that the above condition is implied by $|D| <  (k^2 + 1)2^{k^2\bar{\entropy}_{m_i}}$.
\end{proof}

\subsection{The importance of subgraph isomorphism - Theorem 2}\label{sec:isomorphism_compression}

\begin{customthm}{2}\label{thrm:isomorphism_compression}
Let $p$ be a graph distribution that is invariant to isomorphism, i.e., $p(G') = p(G) \text{ if } G \cong G$. Moreover, consider any algorithm that partitions $G$ in $b$ subgraphs of $k$ vertices. Denote by $\length_{\textnormal{PnC-G}}$ and $\length_{\textnormal{PnC-S}}$ the description length of a PnC compressor that uses a dictionary of atoms encoded as \textit{labelled} graphs and  as isomorphism classes, respectively. The following holds:
\begin{align}
    \E_{G \sim p}[\length_{\textnormal{PnC-S}}(G)]\ \approx  \E_{G \sim p}[\length_{\textnormal{PnC-G}}(G)] -  n (1-\delta)\log k
\end{align}
under the condition that almost all graphs in the dictionary are rigid\footnote{A rigid graph has only the trivial automorphism.}. 
\end{customthm}
Importantly, the compression gains implied by the theorem hold independently of the size of the dictionary, applying e.g., also when the dictionary is equal to the universe and contains all graphs of size $k$ (which amounts to the traditional partitioning baselines).

\begin{proof}
Let $D_G$ be a dictionary of \textit{labelled} graphs of $k$ vertices, i.e. graphs whose vertices are ordered, and $D_S$ the corresponding dictionary of \textit{unlabelled} graphs, i.e., where the atoms are isomorphism classes.

In the context of this comparison we are only interested in the description length of the dictionary subgraphs. For simplicity we will assume that these are encoded with a \textit{categorical} distribution instead of multinomial:
\begin{align*}
    \E_{G\sim p}[\length(\cH_{\text{dict}}|b_{\text{dict}}, D)] 
    &= \E_{G\sim p}\bigg[- \sum_{a\in D}b_a\log q(a)\bigg] 
    =   - b(1-\delta)\sum_{a\in D}q(a)\log q(a) 
    = \frac{n}{k}(1-\delta) \mathbb{H}(D) 
\end{align*}

Hence, in order to compare the two variants, we are interested in the entropy $\mathbb{H}(D)$, which requires enumerating the possible outcomes of the categorical distribution, i.e., the dictionary atoms. 

Denote with $S_a$ the isomorphism class of an atom $a$, i.e., $S_a = \{a' \in \mathfrak{G}_k | a \cong a'\}$
It is known that the size of each $S_a$ is $|S_a| = \frac{k!}{|\text{Aut}(a)|}$ ~\cite{harary2014graphical}, where $\text{Aut}(a)$ the automorphism group of $a$, i.e., all isomorphisms that map the adjacency matrix onto itself.

Since $p(G)$ is isomorphism invariant, then the same will hold for the subgraphs $H \subseteq G$, i.e. $p(H') = p(H) \text{ if } H \cong H$. Hence, regarding PnC-G, it should hold that for each atom $a$ in $D_G$, then all $a'\cong a$ should be also contained in the dictionary and assigned the same probability, i.e.,  $q_{G}(a) = q_{G}(a')$. Therefore, the corresponding probabilties of isomorphism classes in the context of PnC-S should be as follows: $q_{S}(S_a) = \sum_{a \in S_a} q_{G}(a) = |S_a|q_{G}(a)$.

Then, using a similar argument to \cite{choi2012compression}, we can derive the following for the entropy $\mathbb{H}_G(D)$:
\begin{align*}
    \mathbb{H}_G(D) 
    &= - \sum_{a \in D_G}q_G(a)\log q(a) \\
    &= -\sum_{a \in D_G}\frac{q_S(S_a)}{|S_a|}\log \frac{q_S(S_a)}{|S_a|} \\
    & = -\sum_{S_a \in D_S} \sum_{a\in S_a} \frac{q_S(S_a)}{|S_a|}\log \frac{q_S(S_a)}{|S_a|} \\ 
    &= -\sum_{S_a \in D_S} q_S(S_a)\log \frac{q_S(S_a)}{|S_a|} \\ 
    &=\mathbb{H}_S(D) + \sum_{S_a \in D_S} q_S(S_a)\log |S_a| 
    =\mathbb{H}_S(D) + \log k! - \sum_{S_a \in D_S} q_S(S_a)\log|\text{Aut}(a)|
\end{align*}
At this point we will assume that almost all graphs in the dictionary are rigid, or more precisely we require that
${\sum_{S_a \in D_S} q_S(S_a)\log|\text{Aut}(a)| \approx 0}$, which can be also satisfied when non-rigid dictionary atoms have small probability. In practice, although for very small graphs of up to 4 or 5 vertices, many graphs have non-trivial automorphisms, this condition is easily satisfied for larger $k$ (but still of constant size w.r.t. $n$), that were also considered in practice. Then, the result immediately follows:
\begin{align*}
     \E_{G \sim p}[\length_{\text{PnC-G}}(\cH_{\text{dict}}|b_{\text{dict}}, D)] &\approx 
     \E_{G \sim p}[\length_{\text{PnC-S}}(\cH_{\text{dict}}|b_{\text{dict}}, D)] + \frac{n}{k} (1-\delta) \log k! \Longrightarrow\\
        \E_{G \sim p}[\length_{\text{PnC-S}}(G)] &\approx \E_{G \sim p}[\length_{\text{PnC-G}}(G)] - n (1-\delta) \log k,
\end{align*}
where we used Stirling's approximation $\log k! \approx k\log k$.
\end{proof}

\section{Algorithmic Details}

\subsection{Cut encoding}\label{sec:cut_encoding}
Denote the vertex count of subgraph $H_i$ as $k_i$. Further denote with $\bm{m}_c = \{m_{1,1}, m_{1,2}, \ldots, m_{b-1,b}\}$ the vector containing the number of edges between each subgraph pair $i,j$ and $m_c = \sum_{i<j}^b m_{ij}$. The $b$-partite graph $C$ containing the cuts will be encoded hierarchically, i.e. first we encode the total edge count $m_c$, then the pairwise counts $\bm{m}_c$ and finally, for each subgraph pair, we independently encode the arrangement of the edges. For each of these cases, a uniform encoding is chosen, following the rationale mentioned in Section 3 of the main paper. Hence, calculating the length of the encoding boils down to enumerating possible outcomes: 
\begin{align}\label{eq:cut_encoding}
\begin{split}
\length(C|\cH) &= \length(C, m_c, \bm{m}_c|\cH)=  \length(m_c|\cH) + \length(\bm{m}_c|m_c, \cH) + \length(C|\bm{m}_c, m_c, \cH)\\
    &= \log \big(1 + \sum_{j>i}^b k_{i} k_{j}\big) + \log {b(b-1)/2 + m_c - 1 \choose m_c} + \sum_{j>i}^b\log{k_{i} k_{j} \choose m_{ij}}
\end{split}
\end{align}
We make the following remarks: (a) The encoding is the same regardless of the isomorphism class of the subgraphs, and the only dependence arises from their number, as well as their vertex counts.  (b) Small cuts are prioritised, thus the encoding has an inductive bias towards distinct clusters in the graph.
(c) Our cut encoding bears resemblance to those used in non-parametric Bayesian inference for SBMs (e.g., see the section \ref{baselines} on the baselines and \cite{peixoto2019bayesian} for a detailed analysis of a variety of probabilistic models), although a central difference is that in these works the encodings also take the vertex ordering into account.

\subsection{Baseline Encodings} \label{baselines}
\paragraph{Null models.} The description length of the uniform encoding is equal to ${\length(G) = \log (n_{max}+1) + {n \choose 2}}$ and that of the edge list model is ${\length(G) = \log (n_{max} +1) + \log \Big({n \choose 2} + 1\Big) + m \log {n \choose 2} }$.

\paragraph{Clustering. }The encoding we used for the clustering baselines is optimal under SBM assumptions and is obtained from \cite{peixoto2019bayesian} with small modifications. It consists of the following uniform encodings: number of graph vertices, number of graph edges, number of blocks, number of vertices in each block, number of edges inside each block and between each pair of blocks, and finally the arrangements of intra- and inter-block edges (a detailed explaination for each term can be found in \cite{peixoto2019bayesian} and \cite{peixoto_graph-tool_2014}):
\begin{align}\label{eq:partitioning}
\begin{split}
        \length(G) &= \log (n_{\text{max}} + 1) + \log \bigg(n(n-1)/2 + 1\bigg) + \log(n) + \log {n-1 \choose b-1}\\
        &+ \log {b(b+1)/2 + m - 1 \choose m} 
        + \sum_{i=1}^b\log {{k_i \choose 2} \choose m_i} + \sum_{i<j}^b\log {k_i k_j \choose m_{ij}}
\end{split}
\end{align}

\subsection{Dictionary Learning - Continuous Relaxation}\label{sec:relaxation}
In the following section we will relax the Minimum Description Length objective (Eq. (10) in the main paper) by introducing the fractional membership variables $\hat{\bx}$.

The dictionary description length, Eq. (5), can be trivially rewritten as follows:
\begin{equation}
    \length_{\hat{\bx}}(D) = \sum_{a \in \univ} \hat{x}_a \length_{\text{null}}(a). 
\end{equation}

Regarding the description length of the graphs, the membership variables are the ones that select when a subgraph is encoded as a dictionary atom or when with the help of the null model. Note that this is an additional explanation for the necessity of the the dual encoding: except for giving sufficient freedom to the partioning algorithm to choose non-dictionary atoms, it is also necessary for optimisation. 

The relaxation of the graph description length was done as follows:   ${b_a(\hat{\bx}) =  \hat{x}_a b_a}$, ${b_{\text{dict}}(\hat{\bx}) =  \sum_{a \in \univ}{b_a(\hat{\bx})}}$, and $q_{\phi, \hat{\bx}}(a)  = \frac{\hat{x}_a e^{\phi_a}}{\sum_{a' \in \univ} \hat{x}_{a'} e^{\phi_{a'}}}$. The rest of the components of the graph description length are unaffected from the choice of the dictionary. Now Eq. (6)-(8) can be rewritten as:
\begin{align}
\begin{split}
    &\length_{\phi, \hat{\bx}}(b_{\text{dict}}, b_{\text{null}}) 
     = -\log {b \choose b_{\text{dict}}(\hat{\bx})} - 
     b_{\text{dict}}(\hat{\bx})\log (1 - \delta_\phi) - \Big(b-b_{\text{dict}}(\hat{\bx})\Big)\log(\delta_\phi) - \log q_\phi(b)\\
     &\length_{\phi, \hat{\bx}}(\cH_D|b_{\text{dict}}, D) 
     = -\log \Big(b_{\text{dict}}(\hat{\bx})!\Big) +\sum_{a\in \univ}\log \Big(b_a(\hat{\bx})!\Big) - \sum_{a\in \univ}b_a(\hat{\bx}) \log q_{\phi,\hat{\bx}}(a)\\
     &\length_{\text{null}, \hat{\bx}}(\cH_{\text{null}}|b_{\text{null}}, D)
     = -\sum_{H \in \cH}\log q_{\text{null}}(H)(1-\hat{x}_H), \text{ where }          
\  \hat{x}_H = \left\{
\begin{array}{ll}
      \hat{x}_i & \ \exists \ a_i \in \univ \text{ s.t. }H \cong a_i  \\
      0 & \text{otherwise}.\\
\end{array} 
\right. 
\end{split}
\end{align}

To obtain a continuous version of the terms where factorials are involved we used the $\Gamma$ function, where $\Gamma(n+1) = n!$, for positive integers $n$. The rest of the terms are differentiable w.r.t $\hat{\bx}$.

\subsection{Learning to Partition}\label{sec:neural_part}

We remind that our algorithm is based on a double iterative procedure: the external iteration refers to subgraph selection and the internal to vertex selection. 
In order for the algorithm to be able to make decisions, we maintain a representation of two states: the \textit{subgraph state $S^H_t = \{H_1, H_2, \ldots H_t\}$} that summarises the decisions made at the subgraph level (external iteration) up to step $t$ , and the \textit{vertex state $S^{V}_i = \{v_{t_1}, v_{t_2}, \ldots v_{t_i}\}$} that summarises the decisions made at the vertex level (internal iteration) up to the i-th vertex selection. Overall, we need to calculate the probability of $S_T$, where $T$ is the number of iterations:
\begin{align}
\begin{split}
        p_\theta(S_T|G) &= p_\theta(H_T|S^H_{T-1}, G)p_\theta(S^H_{T-1}|G) = \prod_{t=1}^T p_\theta(H_t|S^H_{t-1}, G)\\
        &= \prod_{t=1}^T \bigg(\prod_{i=1}^{k_t} p_\theta(v|S^{V}_{i-1}, k_t, S^H_{t-1}, G)\bigg)p_\theta(k_t|S^H_{t-1}, G)\\
\end{split}
\end{align}

Hence, the parametrisation of the algorithm boils down to defining the vertex count probability $p_\theta(k_t|S^H_{t-1}, G)$ and the vertex selection probability $p_\theta(v|S^{V}_{i-1}, k_t, S^H_{t-1}, G)$, where $v \in V_t$ and $V_t$ the set of the remaining vertices at the step $t$. 

Now we explain in detail how we parametrise each term. First, we use a Graph Neural Network (GNN) to embed each vertex into a vector representation ${\bh(v) = \text{GNN}_v(G)}$, while the graph itself is embedded in a similar way ${\bh(G) = \text{GNN}_G(G)}$. 

Each subgraph is represented by a permutation invariant function on the embeddings of its vertices, i.e., $\bh(H_t) = \text{DeepSets}(\{\bh(v)|v \in H_t\})$, where we used DeepSets \cite{Zaheer2017Deepsets} as a set function approximator. Similarly, the subgraph state summarises the subgraph representations in a permutation invariant manner to ensure that future decision of the algorithm do not depend on the order of the past ones: 
$\bh(S^H_t) = \text{DeepSets}(\{\bh(H_t)|H_t \in S^H_t\})$.

Given the above, the probability of the vertex count at step $t$ is calculated as follows:
\begin{equation}
    p_\theta(k_t|S^H_{t-1}, G) = \text{softmax}_{k_t=1}^{|V_t|}\text{MLP}\big(\bh(S^H_{t-1}), \bh(G)\big),
\end{equation}
where $\text{MLP}$ is Multi-layer perceptron.

As mentioned above, the probability of the selection of each vertex is computed in a way that guarantees connectivity:
\begin{equation}
p_\theta(v|S^{V}_{i-1}, k_t, S^H_{t-1}, G) = \left\{
\begin{array}{ll}
      \text{softmax}_{v \in V_t}\text{MLP}\big(\bh(S^H_{t-1}), \bh(v)\big) &  i=0, v \in V_t \\
     \text{softmax}_{v \in V_t\cap \cN(S^{V}_{i-1})}\text{MLP}\big(\bh(S^H_{t-1}), \bh(v)\big) &  0<i<k_t, v \in V_t\cap \cN(S^{V}_{i-1}) \\
      0 & \text{otherwise},\\
\end{array} 
\right. 
\end{equation}
where $\cN(S^{V}_{i-1})$ denotes the union of the neighbourhoods of the already selected vertices, excluding themselves: $\cN(S^{V}_{i-1}) = \bigcup_{i'=1}^{i-1} \cN(v_{t_{i'}}) - \{v_{t_{1}}, v_{t_{2}}, \ldots v_{t_{i-1}}\}$. Overall, the parameter set $\theta$ is the set of the parameters of the neural networks involved, i.e., GNNs, DeepSets and MLPs.

In the algorithm \ref{algo:partioning} we schematically illustrate the different steps described above.

\begin{algorithm}[h]\label{algo:partioning}
 \caption{Partitioning algorithm}
\SetAlgoLined
 \textbf{Input}: graph $G$\\
 \textbf{Output}: partition $\mathcal{H}$\\
 \textbf{Initialisations:}  $\bh(v) = \text{GNN}_v(G)$,  $\bh(G) = \text{GNN}_G(G)$, $V_1 = V$, $S^H_0 = \emptyset$\\
 $t \leftarrow 1$\\
 \While {$V_t \neq \emptyset$}{
    $k_t \sim p_\theta(k_t|S^H_{t-1}, G)$ \tcp{sample maximum vertex count}
    Initialise $S^{V}_0 = \emptyset$\\
    \While{$i=1\leq k_t \text{ and } \cN(S^{V}_{i-1}) \neq \emptyset $ }{ 
     $v_{t_i} \sim p_\theta(v|S^{V}_{i-1}, k_t, S^H_{t-1}, G)$ \tcp{sample new vertex}
     $S^{V}_i = S^{V}_{i-1} \cup \{v_{t_i}\} $\\
    }
    $H_t = S^{V}_i$\\
    $S^{H}_t = S^{H}_{t-1} \cup \{H_{t}\} $\\
    $ t \leftarrow t + 1 $\\
    }
$\cH = S^{H}_t$
\end{algorithm}

\paragraph{Limitations.} Below we list two limitations of the learnable partitioning algorithm that we would like to address in future work. First, it is well known that GNNs have limited expressivity which is bounded by the Weisfeiler Leman test \cite{xu2018how, morris2019weisfeiler}. The most important implication of this is that they have difficulties in detecting and counting substructures \cite{chen2020can}. Since in our case subgraph detection is crucial in order to be able to partition the graph into repetitive substructures, the expressivity of the GNN might be an issue. Although iterative sampling may mitigate this behaviour up to a certain extent, the GNN will not be able to express arbitrary randomised algorithms.
Modern architectures such as \cite{vignac2020building, bouritsas2020improving} might be more suitable for this task, which makes them good candidates for future exploration on the problem.

Second, more sophisticated inference schemes should be explored, since currently a partition is decoded from the randomised algorithm by taking a single sample from the learned distribution. In particular, currently at each step $t$ the algorithm can only sample $k_t$ vertices as dictated by the initial sampling on the vertex count. However,  there might be benefit from expanding the subgraph more, or stopping earlier than $k_t$ when no other vertex addition can contribute towards a smaller description length. However, there is no control on the stopping criterion apart from the initial vertex count prediction. To this end, it is of interest to explore alternatives that will allow the algorithm to choose from a pool of candidate decisions based on the resulting description lengths (e.g., in hindsight). Further inspiration can be taken from a variety of clustering and graph partitioning algorithms, e.g. k-means or soft clustering in a latent space \cite{DBLP:conf/nips/WilderEDT19, DBLP:conf/nips/LocatelloWUMHUD20}, agglomerative \cite{karger1993global, blondel2008fast} and Markov Chain Monte Carlo as in \cite{peixoto2013parsimonious} where a modified Metropolis-Hastings algorithm is proposed.

\subsection{Special cases of note}

A pertinent question is whether one can determine the optimal way to partition a graph when minimising~\eqref{eq:MDL_general}. Though a rigorous statement is beyond our current understanding, in the following we discuss two special cases:

\textit{(a) Small predefined universe.} When the subgraphs are chosen from a small and predefined $\mathfrak{U}$, one may attempt to identify all the possible atom appearances in $G$ by repeatedly calling a subgraph isomorphism subroutine. The minimisation of~\eqref{eq:MDL_general} then simplifies to that of selecting a subset of subgraphs that have no common edges (as per the definition in Section~\ref{sec:decomp}). The latter problem can be cast as a discrete optimisation problem under an independent set constraint (by building an auxiliary graph the vertices of which are candidate subgraphs and two vertices are connected by an edge when two subgraphs overlap, and looking for an independent set that minimises the description length). 
    
\textit{(b) Unconstrained universe.} When $\univ$ contains all possible graphs, the problem can be seen as a special graph partitioning problem. However, contrary to traditional clustering algorithms \cite{ng2001spectral, karypis1998fast, blondel2008fast, karger1993global}, our objective is not necessary optimised by finding small cuts between clusters (see Appendix \ref{sec:theory}).
    
Since most independent set and partitioning problems are NP-hard, we suspect that similar arguments can be put forward for~\eqref{eq:MDL_general}. 
This highlights the need to design learnable alternatives that can provide fast solutions without the need to be adapted to unseen data. 

\section{Additional Experiments}

\subsection{Ablation studies}

\paragraph{Compression of unseen data.} In the Tables \ref{tab:train_test_molecules} and \ref{tab:train_test_social} we report the compression rates (in bpe) of the training and the test data separately for all the PnC variants. As can be seen, in most of the cases the generalisation gap is small, which implies that there was no evidence of overfitting and the compressor can be used to unseen data with small degradation in the compression quality. 
\begin{table}[h]
\centering
\caption{Average negative log likelihood of train and test data in bpe (molecular distributions).}
\label{tab:train_test_molecules}
\resizebox{\textwidth}{!}{%
\begin{tabular}{@{}lllllll@{}}
\toprule
Dataset   name       & \multicolumn{2}{c}{MUTAG} & \multicolumn{2}{c}{PTC} & \multicolumn{2}{c}{ZINC} \\ \midrule
          Set       & train       & test        & train      & test       & train       & test       \\ \midrule
PnC + SBM         & 3.81        & 3.85        & 4.40       & 4.25       & 3.33        & 3.41       \\
PnC + Louvain     & 2.18        & 2.39        & 2.67       & 2.74       & 1.96        & 1.97       \\
PnC + PropClust   & 2.37        & 2.89        & 3.33       & 3.83       & 2.19        & 2.27       \\
PnC + Neural Part & 2.16±0.02   & 2.28±0.03   & 2.64±0.24  & 2.59±0.21  & 2.01±0.02   & 2.03±0.01  \\ \bottomrule
\end{tabular}%
}
\end{table}
\begin{table}[h]
\caption{Average negative log likelihood of train and test data in bpe (proteins and social network distributions).}
\label{tab:train_test_social}
\centering
\resizebox{\textwidth}{!}{%
\begin{tabular}{@{}lllllll@{}}
\toprule
Dataset name           & \multicolumn{2}{c}{PROTEINS} & \multicolumn{2}{c}{IMDB-B} & \multicolumn{2}{c}{IMDB-M} \\ \midrule
         Set         & train         & test         & train          & test          & train         & test          \\ \midrule
PnC + SBM         & 3.24          & 3.46         & 0.48           & 0.50          & 0.35          & 0.31          \\
PnC + Louvain     & 3.33          & 3.47         & 0.94           & 0.95          & 0.66          & 0.59          \\
PnC + PropClust   & 3.41          & 3.53         & 1.43           & 1.61          & 0.95          & 0.75          \\
PnC + Neural Part & 3.33±0.24     & 3.36±0.29    & 1.01±0.05      & 0.99±0.04     & 0.70±0.03     & 0.63±0.02     \\ \bottomrule
\end{tabular}%
}
\end{table}

\paragraph{Out-of-distribution compression. } In the following experiment we tested the ability of PnC to compress data sampled from different distributions. In particular we trained the Neural Partitioning variant on one of the MUTAG and IMDB-B datasets, and then used the pretrained compressor on the remaining ones. In Table \ref{tab:ood_delta} (left) we report the data as well as the total (data + model) description length, in accordance with the experiments of the main paper. We make the following two observations: (1) As expected, PnC can generalise to similar distributions relatively well (in the table we highlight the MUTAG $\rightarrow$ ZINC and the IMDB-B $\rightarrow$ IMDB-M transfer), but fails to do so when there is significant distribution shift. (2) Although MUTAG contains only approximately 100 graphs, it sufficient to train a compressor that can generalise to a significantly larger dataset (ZINC contains approximately 10K graphs), which is an indication that PnC is sample efficient.

\begin{table}[h]
\centering
\caption{Out of distribution compression (left) and probability of a subgrpaph to belong in the dictionary (right).}
\label{tab:ood_delta}
\resizebox{0.62\textwidth}{!}{%
\begin{tabular}{lllllll}
\toprule
& \multicolumn{6}{c}{Training dataset}                                                              \\ \midrule
           & \multicolumn{2}{c}{MUTAG}     & \multicolumn{2}{c}{IMDB-B} & \multicolumn{2}{c}{same dataset} \\ \cline{2-7} 
Test dataset &
  \multicolumn{1}{c}{data} &
  \multicolumn{1}{c}{total} &
  \multicolumn{1}{c}{data} &
  \multicolumn{1}{c}{total} &
  \multicolumn{1}{c}{data} &
  \multicolumn{1}{c}{total} \\ \midrule
MUTAG      & -             & -             & 6.68           & 7.61          & 2.17±0.02       & 2.45±0.02      \\
PTC        & 4.14          & 4.48          & 8.16           & 8.55          & 2.63±0.26       & 2.97±0.14      \\
ZINC       & \textbf{2.62} & \textbf{2.63} & 6.92           & 6.94          & 2.01±0.02       & 2.07±0.03      \\
PROTEINS   & 4.74          & 4.87          & 4.31           & 4.44          & 3.34±0.25       & 3.51±0.23      \\
IMDB-B & 1.83          & 1.86          & -              & -             & 1.00±0.04       & 1.05±0.04      \\
IMDB-M  & 1.37          & 1.39          & \textbf{0.74}  & \textbf{0.77} & 0.66±0.05       & 0.72±0.05    \\
\bottomrule
\end{tabular}%
}
\quad
\resizebox{0.25\textwidth}{!}{%
\begin{tabular}{ll}
\toprule
dataset    & 1 - $\delta$  \\ \midrule
MUTAG      & 0.998 \\
PTC        & 0.995 \\
ZINC       & 0.999 \\
PROTEINS   & 0.999 \\
IMDB-B & 0.995 \\
IMDB-M  & 0.997 \\
\bottomrule
\end{tabular}%
}
\end{table}

\paragraph{How frequently do we encounter dictionary subgraphs?} In Table \ref{tab:ood_delta} (right), we report the probability $1-\delta$  for the Neural Partitioning variant of PnC, i.e., the estimated probability of an arbitrary subgraph to belong in the dictionary. Interestingly, since the values are very close to 1, it becomes evident that the partitioning algorithm learns to detect frequent subgraphs in the distribution, which (following Theorem 1b) can in part justify the high compression gains of PnC in all the datasets.

\subsection{Reducing the model size of deep generative models}\label{sec:ablation_likelihood}
\subsubsection{Smaller architectures}
It is made clear by the experimental results of section \ref{sec:main_evaluation} that deep neural compression is particularly costly due to heavy overparametrisation. Yet, we also observe that these models achieve strong results in terms of likelihood. Is it possible to strike a better balance between number of parameters and compression cost for a deep generative model?
To investigate this, we have conducted the following experiment. We trained 5 GRAN models that differ in parameter count, on 3 different datasets, and monitored the total BPE. 

\begin{figure}[t]
        \centering
         \includegraphics[width=\linewidth]{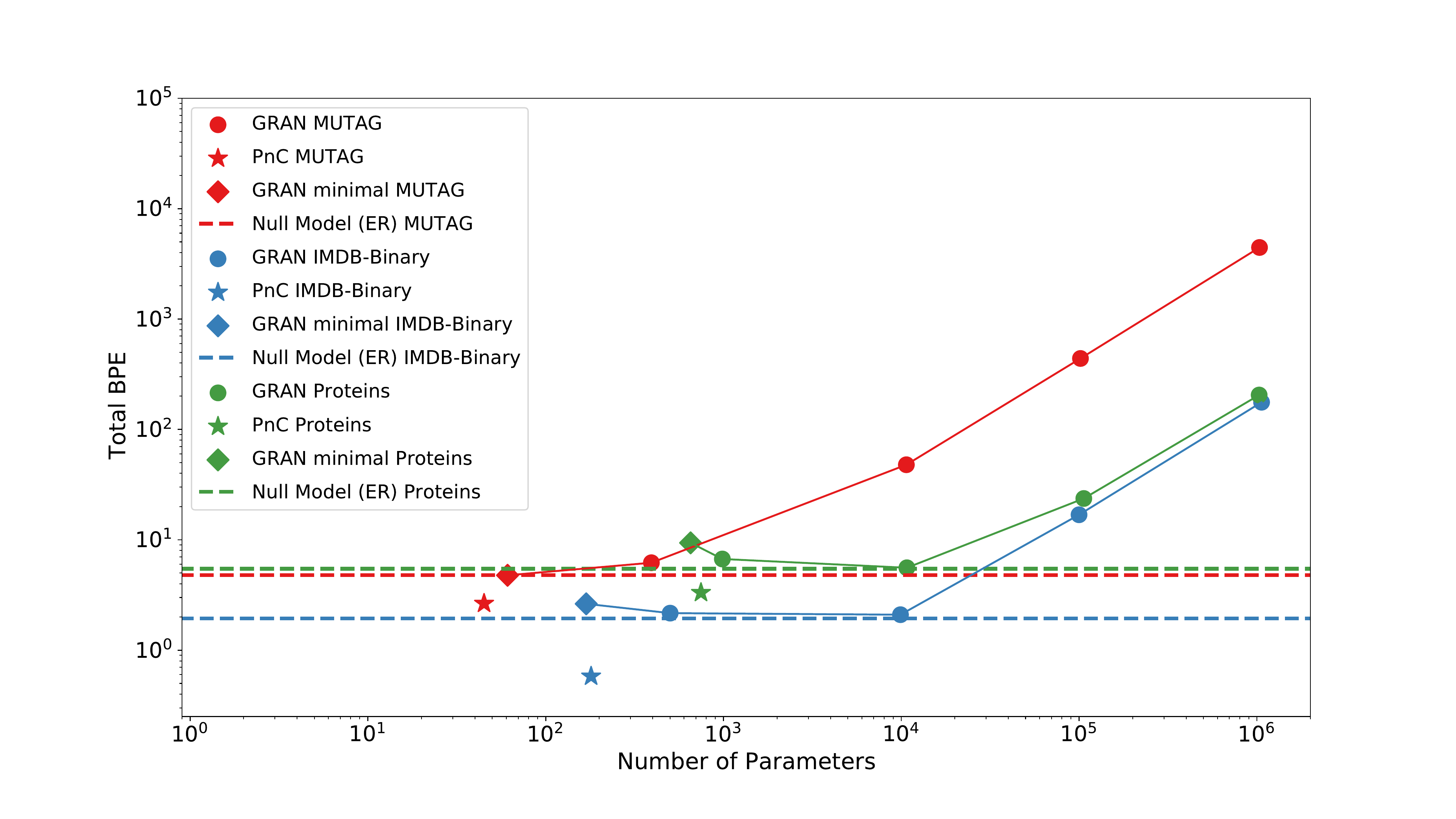}
        \label{fig:ablation_plot} 
    \caption{Total BPE of likelihood-based models as a function of the parameter count. \textit{GRAN minimal} refers to the minimum working GRAN model that does not feature attention and multiple mixture components. The non-parametric ER model is represented with dashed lines.\vspace{-4mm}}
    \label{fig:ablation_likelihood.}
\end{figure}

In order to consistently scale the number of parameters across these 5 different models, we have fixed the GNN depth for all models to one, and set for each model the size of the embedding, attention, and hidden dimension, to a constant $c$. Using a different $c$ for each model allows us to explore different scales for the parameter count of the GRAN model. 
Furthermore, to facilitate comparison with PnC, one of the 5 models is trained without attention and features a reduced amount of mixture components. This is the minimum, in terms of parameter count, working  instantiation of GRAN. Finally, we also considered the BPE for the null Erd\H{o}s-Renyi (ER) model. 
Figure \ref{fig:ablation_likelihood.} plots the total BPE of the different models against the number of parameters.

\paragraph{Results.} 
In the low parameter regime, the GRAN models are not capable of outperforming the null models and fall significantly behind PnC. At scales that range from $10^3$ to $10^4$ parameters, we observe slight improvements in the total BPE of GRAN on the Proteins and the IMDB-Binary datasets. However, the improved likelihood is not able to compensate sufficiently for the increase in the number of parameters. This becomes more pronounced on larger scales, where GRAN experiences diminishing returns as the cost of parameters outpaces the likelihood gains.
On the other hand, the results consistently worsen as the number of parameters grows on MUTAG. In this case, the size of the dataset is an additional detrimental factor that weighs against overparametrised models.
Overall, the experiment suggests that, as the number of parameter grows, the off-the-shelf GRAN model becomes increasingly inefficient and is thus not well suited for compression. 

\subsubsection{Pruning}\label{sec:pruning}
Based on the results of the previous section, parameter search alone cannot mitigate the cost of overparametrisation. A more efficient approach to manage the tradeoff between model size and data likelihood is required. However, as shown on table \ref{tab:ratios}, a (at least) two order of magnitude reduction in the model size without a decrease in the data likelihood is required for a more competitive neural compressor with deep generative models. In the following section, we experimented with a combination of modern model compression techniques as a heuristic to reduce the model size. 

Model compression techniques aim to reduce the size of a given model while maintaining its performance.  Research in model compression has empirically demonstrated that large models can often be considerably shrunk without suffering from major performance losses. Combined with parameter search and mixed precision training, model compression may result in more cost-effective neural compressors. We investigate the feasibility of such an approach in the following experiment.

\textit{First, we decrease the model size by hyper-parameter search.} We fix the depth of both GraphRNN and GRAN to the one provided in the original implementations, and gradually reduce their width to identify a compact version of the network that maintains high performance. This leads to fixing the width of both GraphRNN and GRAN to 16. \textit{Then, we train both models using an iterative weight pruning technique.} We opt for global unstructured L1 weight pruning, using the lottery ticket procedure \cite{DBLP:conf/iclr/FrankleC19} that has been shown to be effective in the literature. The method we utilised proceeds in the following way: A model is trained for $T$ iterations ($T$ is a hyperparameter chosen based on the convergence and running time of the models on each dataset), then a percentage of the weights are pruned (25\% in our case). After pruning, the unpruned weights are reset to their initial state and the process is repeated from the beginning using the new pruned network. This yields up to a 10x reduction in the size of the model in most datasets (we report the best total description length between all pruning phases). \textit{Finally, we attempted to reduce the storage size of the model weights using half precision}. Traditionally, NNs are trained with 32-bit floating point numbers. Recently, progress has been made in mixed precision training which can enable the use of 16-bit tensors \cite{MicikeviciusNAD18}. We follow the same procedure and at the end of training we store the model weights using 16 bits.

Tables \ref{tab:single_vs_half_mol} and \ref{tab:single_vs_half_social} contain the results of both single and half precision pruned models on all datasets. As it can be observed in the results, both models benefit significantly from this hybrid approach, albeit at the cost of reduced data likelihoods. \textit{However, PnC is still able to outperform the pruned versions}. Although our approach to model compression is by no means exhaustive, it becomes evident that the procedure is quite tedious and choosing the right trade-off between the data likelihood and the model size is based on heuristics, hence the minimisation of the total description length cannot be guaranteed. Nevertheless, we believe that this an important research direction that should be further explored in a more principled manner.

\begin{table}[h]
\centering
\caption{Minimum model compression ratios required for overpametrised neural compressors to outperform PnC. We assume  zero degradation of the data likelihood.}
\label{tab:ratios} 
\begin{tabular}{@{}lll@{}}
\textbf{dataset} & \textbf{GraphRNN} & \textbf{GRAN} \\ \midrule
MUTAG            & x2980           & x7657        \\
PTC              & x995              & x6668         \\
ZINC             & x112              & x227          \\
PROTEINS         & x105               & x303          \\
IMDBB            & infeasible        & x1745          \\
IMDBM            & infeasible        & x2262
\end{tabular}
\end{table}

\begin{table}[h]
\caption{Pruning deep graph generators with single and half precision (molecular distributions)}
\label{tab:single_vs_half_mol}
\resizebox{\textwidth}{!}{%
\begin{tabular}{@{}llcccccccc@{}}
\toprule
                       dataset name             & \multicolumn{3}{c}{MUTAG}        & \multicolumn{3}{c}{PTC}       & \multicolumn{3}{c}{ZINC}          \\
\midrule                      &                          data & total            & params & data       & total   & params & data & total             & params \\ \midrule
            GraphRNN (half)             
        & 4.70 & 10.77  & 1.08K
        & 9.53 & 12.10 &  1.10K
        & 3.89 & 4.10 &  2.64K \\
        GRAN (half)             & 2.41&14.84 & 2.21K       & 4.35 & 9.86& 2.36K    
        & 3.26 & 3.38 & 1.67K \\
   GraphRNN (single) & 1.95 & 12.39 & 1.08K & 2.16 & 6.71 & 1.10K & 1.79 & {2.02} & 1.90K\\
   
    GRAN (single) & 2.59  & 24.56 & 2.23K & 4.31 &14.00 &2.36K &3.26 & 3.47 & 1.69K \\
 \bottomrule%
\end{tabular}%
}
\end{table}

\begin{table}[h]
\caption{Pruning deep graph generators with single and half precision (proteins and social network distributions)}
\label{tab:single_vs_half_social}
\resizebox{\textwidth}{!}{%
\begin{tabular}{@{}llcccccccc@{}}
\toprule
                       dataset name             & \multicolumn{3}{c}{PROTEINS}        & \multicolumn{3}{c}{IMDB-B}       & \multicolumn{3}{c}{IMDB-M}          \\
\midrule                      &                          data & total            & params & data       & total   & params & data & total             & params \\ \midrule
            GraphRNN (half)             
        & 27.10 & 27.47 & 1.43K
        & 4.21 & 4.49  & 1.28K
        & 2.91 & 3.16 & 1.20K \\
        GRAN (half)             &3.89 & 4.70 & 3.16K       & 0.89& 1.41 &2.39K    
        &0.61 & 1.10 & 2.31K \\
    GraphRNN (single) & 2.63 & 3.76 & 2.56K &1.43 & 1.92& 1.28K & 0.91 & 1.39 & 1.28k\\
   GRAN (single) & 4.28 & 5.11 & 1.78K & 0.84 & 1.75 & 2.38K & 0.55 & 1.41 &2.31K \\
 \bottomrule%
\end{tabular}%
}
\end{table}

\subsection{Vertex and Edge attributes}\label{sec:attrs}
Our method can be easily extended to account for the presence of discrete vertex and edge attributes, the distribution of which can also be learned from the data. Assuming a vertex attribute domain $A_V$ and an edge attribute domain $A_E$, we can use the following simple encodings for a graph with $n$ vertices and $m$ edges:
\begin{equation}
    \length(\bX_V) = n\log |A_V| \text{ and } \length(\bX_E) = m\log |A_E|,
\end{equation}
where $\bX_V \in \mathbb{N}^{|V|\times |A_V|}$ the vertex attributes  and $\bX_E \in \mathbb{N}^{|E|\times |A_E|}$ the edge attributes. One could also choose a more sophisticated encoding by explicitly learning the probability of each attribute. 

In this case, the dictionary becomes even more relevant, since when simply partitioning the graph, the attributes will still have to be stored in the same manner for each subgraph and each edge in the cut. Hence, in the absence of the dictionary it will be impossible to compress the attributes.

In Table \ref{tab:attributes}, we showcase a proof of concept in the attributed MUTAG and PTC MR datasets, which are variations of those used for structure-only compression in the main paper. Vertex attributes represent atom types and edge attributes represent the type of the bond between two atoms. As mentioned in the previous paragraph, it is clear that non-dictionary methods are hardly improving w.r.t the null model, which is mainly due to the fact that the attributes constitute the largest portion of the total description length. Another interesting observation is that since the clustering algorithms we used are oblivious to the existence of attributes, they are less likely to partition the graph in such a way that the attributed subgraphs will be repetitive, unless structure is strongly correlated with the attributes. This becomes clear in the PTC MR dataset, where between the different PnC variants, the neural partitioning performs considerably better, since the partitioning is optimised in coordination with the dictionary. In Figure \ref{fig:substructures.} of the main paper, we show the most probable substructures that the Neural Partioning yields for the MUTAG dataset. It is interesting to observe that typical molecular substructures are extracted. This highlights an interesting application of molecular graph compression, i.e., discovering representative patterns of the molecular distribution in question.

\begin{table}[h!]
\centering
\caption{Experimental evaluation on the attributed MUTAG and PTC MR molecular datasets. \textcolor{red}{\textbf{First}}, \textcolor{violet}{\textbf{Second}}, \textcolor{blue}{\textbf{Third}}}
\label{tab:attributes}
{\small
\resizebox{\textwidth}{!}{%
\begin{tabular}{@{}llcccccc@{}}
\toprule
\multicolumn{1}{c}{} &
  Dataset name &
  \multicolumn{3}{c}{Atrributed MUTAG} &
  \multicolumn{3}{l}{Atrributed PTC MR} \\ \cmidrule(l){2-8} 
\multicolumn{1}{c}{\multirow{-2}{*}{\begin{tabular}[c]{@{}c@{}}Method\\ Family\end{tabular}}} &
   &
  data &
  total &
  params &
  data &
  total &
  params \\ \midrule
                       & Uniform (raw adjacency) & -    & 13.33                       & -  & -    & 16.32 & -   \\
                       & Edge list               & -    & 12.62                       & -  & -    & 14.06 & -   \\
\multirow{-3}{*}{Null} & Erd\H{o}s-Renyi             & -    & 9.38                        & -  & -    & 10.87 & -   \\ \midrule
Clustering             & SBM-Bayes               & -    & 9.17                        & -  & -    & 10.61 & -   \\
                       & Louvain                 & -    & 9.37                        & -  & -    & 10.76 & -   \\
                       & PropClust               & -    & 9.52                        &    & -    & 10.80 &     \\ \midrule
PnC &
  PnC + SBM &
  6.56 &
  7.49 &
  78 &
  8.05 &
  {\color{blue} \textbf{9.49}} &
  158 \\
 &
 
  PnC + Louvain &
  3.52 &
  {\color{red} \textbf{4.45}} &
  78 &
  5.56 &
  {\color{violet} \textbf{7.65}} &
  200 \\
  
    & PnC + PropClust   
    & 5.21 & {\color{blue} \textbf{6.30}} & 54 & 8.51 & 9.58  & 118 \\ \midrule
 &
  PnC + Neural Part. &
  3.83±0.06 &
  {\color{violet} \textbf{4.78±0.12}} &
  74±6 &
  5.19±0.39 &
  {\color{red} \textbf{6.49±0.54}} &
  170±30 \\ 
  \bottomrule
\end{tabular}%
}
}
\end{table}

\section{Implementation Details}\label{sec:implementation}
\paragraph{Datasets:} We evaluated our method on a variety of datasets that are well-established in the GNN literature. In specific, we chose the following from the TUDataset collection \cite{Morris+2020}: the molecular datasets MUTAG \cite{debnath1991structure, DBLP:conf/icml/KriegeM12} (mutagenicity prediction) and PTC-MR \cite{helma2001predictive, DBLP:conf/icml/KriegeM12} (carcinogenicity prediction), the protein dataset PROTEINS \cite{borgwardt2005protein,dobson2003distinguishing} (protein function prediction - vertices represent secondary structure elements and edges either neighbourhoods in the aminoacid sequence or proximity in the 3D space) and the social network datasets IMDBBINARY and IMDBMULTI \cite{yanardag2015deep} (movie collaboration datasets where each graph is an ego-net for an actor/actress). We also experimented with the ZINC dataset \cite{irwin2012zinc, DBLP:conf/icml/KusnerPH17,gomez2018automatic, DBLP:conf/icml/JinBJ18} (molecular property prediction), which is a larger molecular dataset from the dataset collection introduced in \cite{dwivedi2020benchmarking}. A random split is chosen for the TUDatasets (90\% train, 10\% test), since we are not interested in the class labels, while for the ZINC dataset we use the split given by the authors of \cite{dwivedi2020benchmarking} (we unify the test and the validation split, since we do not use the validation set for hyperparameter tuning/model selection).

\paragraph{PnC model architecture and hyperparameter tuning:} The GNN used for the Neural partitioning variant of PnC is a traditional Message Passing Neural Network \cite{gilmer2017neural}, where a general formulation is employed for the message and the update functions (i.e., we use Multi-layer Perceptrons similar to \cite{Loukas2020What}). We optimise the following hyperparameters: batch size in \{16, 64, 128\}, network width in \{16, 64\} number of layers in \{2, 4\}.
The learning rate for the updates of the dictionary and probabilistic model parameters was 1 and 0.1 for the fixed partitioning and the neural partitioning variants respectively, while the learning rate of the GNN (neural partitioning only) was set to 0.001. For all the variants we further tune
the maximum number of vertices for the dictionary atoms $k$ in \{6, 8, 10, 12\}. Note that the last hyperparameter mainly affects the optimisation of the Neural Partitioning variant: small values of $k$ will constrain the possible subgraph choices, but will facilitate the network to find good partitions by exploitation. On the other hand, larger values of $k$ will encourage exploration, but the optimisation landscape becomes significantly more complex, thus in some cases (mainly for social networks, where there is a larger variety of non-isomorphic subgraphs) we observed that the optimisation algorithm could not converge to good solutions. 

We optimise each PnC variant for 100 epochs and report the result on the epoch where the description length of the training set is minimum. The best hyperparameter set is also chosen w.r.t the lowest training set description length, and after its selection, we repeat the experiment for 3 different seeds (in total). Table \ref{tab:hyperparams} shows the chosen hyperparameters.

\begin{table}[h]
\centering
\caption{Chosen hyperparameters for each dataset (PnC + NeuralPart)}
\label{tab:hyperparams}
\begin{tabular}{lllll}
\toprule
dataset    & batch size & width & number of layers & k  \\ \midrule
MUTAG      & 16 & 16 & 2 & 10 \\
PTC        & 16 & 16 & 2 & 10 \\
ZINC       & 128 & 16 & 2 & 10 \\
PROTEINS   & 16 & 16 & 4 & 8 \\
IMDB-B & 16 & 16 & 2 & 8 \\
IMDB-M  & 64 & 64 & 2 & 8 \\
\bottomrule
\end{tabular}%
\end{table}

We implement our framework using PyTorch Geometric \cite{Fey/Lenssen/2019}, while the predefined partitioning algorithms were implemented using graph-tool \cite{peixoto_graph-tool_2014} for the SBM fitting and scikit-network \cite{JMLR:v21:20-412} for the Louvain and the Propagation Clustering algorithms. To track our experiments we used the wandb platform \cite{wandb}.

\paragraph{Deep generative models and pruning.}
For the generative model baselines, we have used the official implementations provided in the corresponding repositories\footnote{ \url{https://github.com/JiaxuanYou/graph-generation} \& \url{https://github.com/lrjconan/GRAN}}. For GraphRNN, we trained with the default parameters provided with the official implementation and only tuned the number of training epochs according to the time required for convergence. For GRAN, we adopt one of the configurations provided in the official repository with minor modifications. Namely, we used a DFS ordering, stride and block size 1, 20 Bernoulli mixture components for the parametrisation of the likelihood, and switched of the subgraph sampling feature.

For our iterative pruning protocol, we fix the same number of pruning iterations for both models on each dataset. Specifically we use $\{450, 270,10, 90, 90, 90\}$ total epochs and a pruning interval $T$ of $\{50, 30, 1, 10, 10, 10\}$ for MUTAG, PTC, ZINC, PROTEINS, IMDB-B, and IMDB-M respectively. We used a 25\% pruning percentage, which lead to a 10-fold reduction in model size in most cases. Further pruning was not found to be consistently beneficial in the parameter ranges that we experimented on.

\paragraph{Model parameter cost. }
For the PnC variants we could seamlessly use half precision (16 bits) to store the model parameters (section 4.3 in the main paper) without sacrificing compression quality. However, as discussed in Appendix \ref{sec:pruning} we were not able to retain similar likelihood estimates when storing with half precision the weights of deep generative models,
hence in the results reported in the main tables we used 32 bits to store the model weights\footnote{In a preliminary version of the paper we assumed a 16-bit encoding of the weights without likelihood losses. However, our subsequent implementation and experimentation with half-precision deep graph generators demonstrated that this might not be possible in practice.}. Additionally, regarding the pruned versions of deep generative models, we need to send the locations of the non-zero weights for each parameter matrix in $\mathbb{R}^{d_1 \times d_2}$, which are encoded as follows: $\log(d_1 d_2 + 1) + \log{d_1 d_2\choose e}$, where $e$ is the number of the non-zero elements. 
For all methods compared, the decompression algorithm and the neural network architectures are assumed to be public, hence they do not need to be transmitted.

\paragraph{Isomorphism. } In order to speed-up isomorphism testing between dictionary atoms and the subgraphs that the partitioning algorithm yields, we make the following design choices: (a) Dictionary atoms are sorted by their frequencies of appearance (these are computed by an exponential moving average that gets updated during training). In this way the expected number of comparisons drops to $O(1)$ from $O(|D|)$. (b) We choose the parameter $k$ to be a small constant value (i.e., does not scale with the number of vertices of the graph), as previously mentioned. It becomes clear, that except for the importance of $k$ in the optimisation procedure, it also plays a crucial role for scalability, since as mentioned in the introduction in the main paper, solving the isomorphism problem quickly becomes inefficient when the number of vertices increases. (c) Additionally, one can chose to approximate isomorphism with faster algorithms, such as the Weisfeiler-Leman test \cite{weisfeilerreduction}, or with more expressive Graph Neural Networks, such as \cite{bouritsas2020improving, vignac2020building, beaini2020directional}. These algorithms, will always provide a correct negative answer whenever two graphs are non-isomorphic, but a positive answer does not always guarantee isomorphism. In that case, exact isomorphism can be employed only when the faster alternatives give a positive answer.

\section{Translating probabilities into codes}
In the following section, we explain how a partitioned graph can be represented into a bitstream using our probabilistic model. The general principle for modern entropy encoders (Arithmetic Coding \cite{witten1987arithmetic} and Asymmetric Numeral Systems \cite{duda2013asymmetric}) is that both the encoder and the decoder need to possess the cumulative distribution function (c.d.f.) of each component they are required to encode/decode. Hence, the encoder initially sends to the decoder the parameters of the model $\phi$ using a fixed precision encoding (e.g., we used 16-bits for our comparisons). 
The rest of the bitstream is described below:
\begin{itemize}
    \item \textbf{Dictionary.} The dictionary is sent as part of the preamble of the message. It consists of the following:
    
    (a) the size of the dictionary (we assume fixed precision for this value),
    
    (b) a sequence of dictionary atoms encoded with the null model, i.e., the message includes the number of vertices, the number of edges and finally the adjacency matrix: $k_i, m_i, E_i$ (see Eq. (2) in the main paper).
    
    \item \textbf{Graphs: } Subsequently, each graph is sequentially transmitted. The message contains the following:
    
    (a) the total number of subgraphs and the number of dictionary subgraphs $b$ and $b_\text{dict}$ that are encoded using the parameters of the categorical distribution $q_\phi(b)$ and the binomial distribution $ \text{Binomial}(b_{\text{dict}}|b; \phi)$. The c.d.f. of the binomial distribution can be computed using a factorisation described in 
    \cite{DBLP:phd/ethos/Steinruecken15},
    
    (b) the subgraphs that belong in the dictionary, which are encoded using the multinomial distribution $q_\phi(\cH_{\text{dict}}|D)$. As above, a factorisation described in \cite{DBLP:phd/ethos/Steinruecken15} can be used to compute the c.d.f.,
    
    (c) the non-dictionary subgraphs. These are encoded with the null model (same with the encoding of dictionary atoms as mentioned above),
    
    (d) the cuts, which are encoded using Eq. \eqref{eq:cut_encoding}.
    
\end{itemize}

Several of our encodings involve uniform distributions over combinations of elements (e.g., for the adjacency matrix in the null model). To compute them, we can either factorise the distribution as in \cite{steinruecken2016compressing} in order to efficiently compute the c.d.f, or use a ranking function (and its inverse for the decoder) that maps a combination to its index in lexicographic order (e.g., see \cite{kreher2020combinatorial}).

\end{document}